\documentclass[10pt,twocolumn,letterpaper]{article}
\usepackage{cvpr}              

\usepackage{graphicx}
\usepackage{amsmath}
\usepackage{amssymb}
\usepackage{booktabs}
\usepackage{multirow}
\usepackage{capt-of}
\usepackage{cuted}
\usepackage{xfrac}
\newcommand{\tabincell}[2]{\begin{tabular}{@{}#1@{}}#2\end{tabular}}
\usepackage[accsupp]{axessibility}

\usepackage[pagebackref,breaklinks,colorlinks]{hyperref}

\usepackage[capitalize]{cleveref}
\crefname{section}{Sec.}{Secs.}
\Crefname{section}{Section}{Sections}
\Crefname{table}{Table}{Tables}
\crefname{table}{Tab.}{Tabs.}

\def\cvprPaperID{1064} 
\def\confName{CVPR}
\def\confYear{2022}

\begin{document}
	
	\title{MAT: Mask-Aware Transformer for Large Hole Image Inpainting}
	
	\author{%
		Wenbo Li\textsuperscript{1} \quad Zhe Lin\textsuperscript{2} \quad Kun Zhou\textsuperscript{3} \quad Lu Qi\textsuperscript{1} \quad Yi Wang\textsuperscript{4}\thanks{Corresponding author} \quad Jiaya Jia\textsuperscript{1} \\
		$^{1}$The Chinese University of Hong Kong \quad ${^2}$Adobe Inc. \\
		${^3}$The Chinese University of Hong Kong (Shenzhen) \quad ${^4}$Shanghai AI Laboratory \\
		{\tt\small \{wenboli,luqi,leojia\}@cse.cuhk.edu.hk} \\
		{\tt\small zlin@adobe.com \quad kunzhou@link.cuhk.edu.cn \quad wangyi@pjlab.org.cn}
		\vspace{-0.4in}
	}
	\maketitle

	\begin{strip}\centering
		\includegraphics[width=1.0\linewidth]{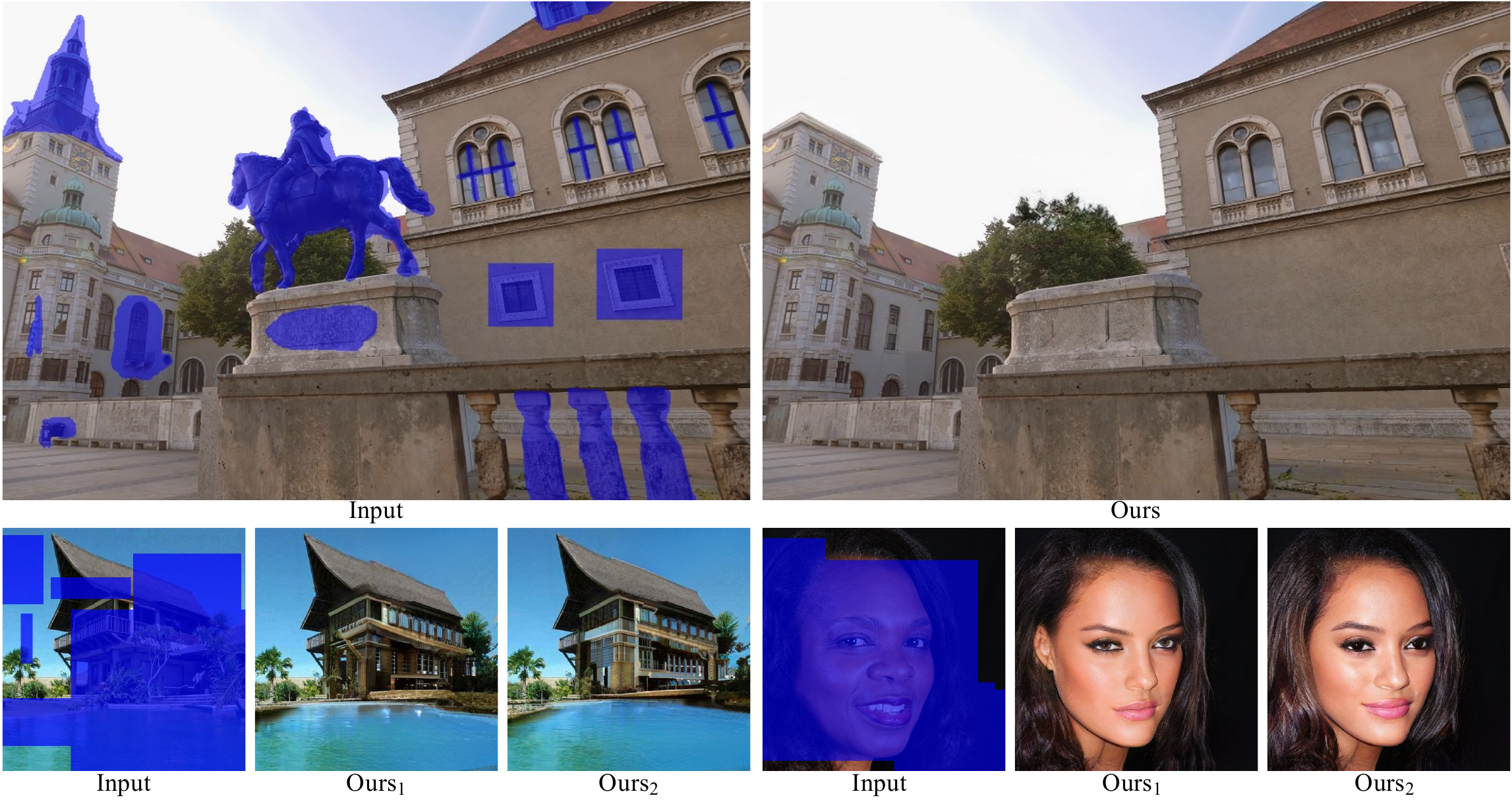}
		\captionof{figure}{The proposed MAT supports photo-realistic and pluralistic large hole image inpainting. The first example is a real-world high-resolution image and the other two examples ($512 \times 512$) are from Places~\cite{zhou2017places} and FFHQ~\cite{karras2019style} datasets. \label{fig:teasing}} 
	\end{strip}
	
	\begin{abstract}

		Recent studies have shown the importance of modeling long-range interactions in the inpainting problem. To achieve this goal, existing approaches exploit either standalone attention techniques or transformers, but usually under a low resolution in consideration of computational cost. In this paper, we present a novel transformer-based model for large hole inpainting, which unifies the merits of transformers and convolutions to efficiently process high-resolution images. We carefully design each component of our framework to guarantee the high fidelity and diversity of recovered images. Specifically, we customize an inpainting-oriented transformer block, where the attention module aggregates non-local information only from partial valid tokens, indicated by a dynamic mask. Extensive experiments demonstrate the state-of-the-art performance of the new model on multiple benchmark datasets. Code is released at \url{https://github.com/fenglinglwb/MAT}.
		
	\end{abstract}
	
	\vspace{-0.1in}
	\section{Introduction}
	\label{sec:intro}
	
	Image completion (a.k.a. inpainting) is a fundamental problem in computer vision, which aims to fill missing regions with plausible contents. It has many applications including image editing~\cite{jo2019sc}, image re-targeting~\cite{cho2017weakly}, photo restoration~\cite{wan2020bringing,wan2020old} and object removal~\cite{barnes2009patchmatch}.
	
	In inpainting, modeling the contextual information is crucial, especially for large masks. Creating reasonable structures and textures for the missing areas demands contextual understanding, using distant information according to non-local priors~\cite{buades2005non,mairal2009non,berman2016non,wang2018non} in images. Previous works employ stacked convolutions to reach large receptive fields and model long-range relationships, which works well on aligned (\eg, faces, bodies) and texture-heavy (\eg, forests, water) data. When processing images with complicated structures (\ie, the first example in the $2_{nd}$ row in Figure~\ref{fig:teasing}), it is difficult for fully convolutional neural networks (CNNs) to characterize the semantic correspondences between distant areas. This is mainly due to the inherent properties of CNNs, the slow growth of the effective receptive field and the inevitable dominance of nearby pixels. To explicitly model long-range dependencies in inpainting,~\cite{yu2018generative,xie2019image,yi2020contextual} propose to employ attention modules in the CNN-based generator. However, limited by the quadratic computational complexity, the attention module is merely applied to relatively small-scale feature maps with a few times, where long-range modeling is not fully exploited.
	
	In contrast to applying attention modules to CNNs, transformer~\cite{vaswani2017attention} is a natural architecture to handle non-local modeling, where attention is a basic component in every block. Recent advances~\cite{wan2021high,zheng2021tfill,yu2021diverse} adopt transformer structures to address the inpainting problem. Nonetheless, affected by the complexity issue, existing works only employ transformers to infer low-resolution predictions (\eg\ $32 \times 32$) for subsequent processing, hence the produced image structure is coarse, compromising the final image quality, especially on large-scale masks.
	
	In this paper, we develop a new inpainting transformer, capable of generating high-resolution completed results for large mask inpainting. Due to the lack of useful information in some regions (this is common when the given mask rules out most pixels), we find the commonly utilized transformer block (LN$\rightarrow$MSA$\rightarrow$LN$\rightarrow$FFN) exhibits inferior performance in adversarial training. In this regard, we customize the vanilla Transformer block to increase optimization stability and also improve performance, by removing the conventional layer normalization~\cite{ba2016layer} and replacing the residual learning with fusion learning using feature concatenation. We analyze why these modifications are crucial for learning and empirically demonstrate they are non-trivial. Also, to handle possible heavy interactions between all tokens extracted from the high-resolution input, we propose a new variant of multi-head self-attention (MSA), named multi-head contextual attention (MCA). It computes non-local relations only using partial valid tokens. The selection of adopted tokens is indicated by a dynamic mask, which is initialized by the input mask and updated with spatial constraints and long-range interactions, improving the efficiency at no cost of effectiveness. Additionally, we incorporate a novel style manipulation module into the proposed framework, inherently supporting pluralistic generation. As shown in Fig.~\ref{fig:teasing}, our method successfully fills large holes with visually realistic and exceptionally diverse contents. Our contributions are summarized as:
	
	\begin{itemize}
		\item We develop a novel inpainting framework MAT. It is the first transformer-based inpainting system capable of directly processing high-resolution images.
		\item We meticulously design components of MAT. The proposed multi-head contextual attention conducts long-range dependency modeling efficiently by exploiting valid tokens, indicated by a dynamic mask. We also propose a modified transformer block to make training large masks more stable. Moreover, we design a novel style manipulation module to improve diversity.
		\item MAT sets new state of the arts on multiple benchmark datasets including Places~\cite{zhou2017places} and CelebA-HQ~\cite{karras2018progressive}. It also enables pluralistic completion.
	\end{itemize}
	
	%
	%
	%
	%
	%
	%
	
	\section{Related Work}
	\label{sec:rela}
	
	Image completion has been a longstanding problem in computer vision. Early diffusion-based methods~\cite{bertalmio2000image, ballester2001filling} propagate neighboring undamaged information to the holes. Within an internal or external searching space, patch-based or exemplar-based approaches~\cite{hays2007scene,sun2005image,criminisi2004region,le2011examplar,criminisi2003object,ding2018image,lee2016laplacian} borrow patches with similar appearance based on human-defined distance metrics to complete missing regions. PatchMatch~\cite{barnes2009patchmatch} proposes a multi-scale patch searching strategy to accelerate the inpainting process. Moreover, partial differential equation~\cite{grossauer2004combined,bertalmio2006strong} and global or local image statistics~\cite{levin2003learning,ghorai2019multiple,fadili2009inpainting} are vastly studied in the literature. Though traditional methods can often obtain visually realistic results, the lack of high-level understanding hinders them from generating semantically reasonable contents.
	
	In the few years, deep learning has achieved great success on the image completion. Pathak \etal~\cite{pathak2016context} bring the adversarial training~\cite{goodfellow2014generative} to inpainting and utilize an encoder-decoder-based architecture to fill holes. Afterwards, numerous variants~\cite{yan2018shift,zeng2019learning,liu2020rethinking,wang2018image} of the U-Net structure~\cite{ronneberger2015u} have been developed for image completion. Besides, more sophisticated networks or learning strategies are proposed to generate high-quality images, including global and local discrimination~\cite{iizuka2017globally}, contextual attention~\cite{yu2018generative,liu2019coherent,xie2019image,yi2020contextual}, partial~\cite{liu2018image} and gated~\cite{yu2019free} convolution, \etc. Multi-stage generation has also received a great amount of attention, where intermediate clues like object edges~\cite{nazeri2019edgeconnect}, foreground contours~\cite{xiong2019foreground}, structures~\cite{ren2019structureflow} and semantic segmentation maps~\cite{song2018spg} are extensively exploited. To allow for high-resolution image inpainting, a few attempts have been made to develop progressively generation systems, such as~\cite{zhang2018semantic,guo2019progressive,li2020recurrent,zeng2020high,oh2019onion}.
	
	\begin{figure*}[t]
		\begin{center}
			\includegraphics[width=0.98\linewidth]{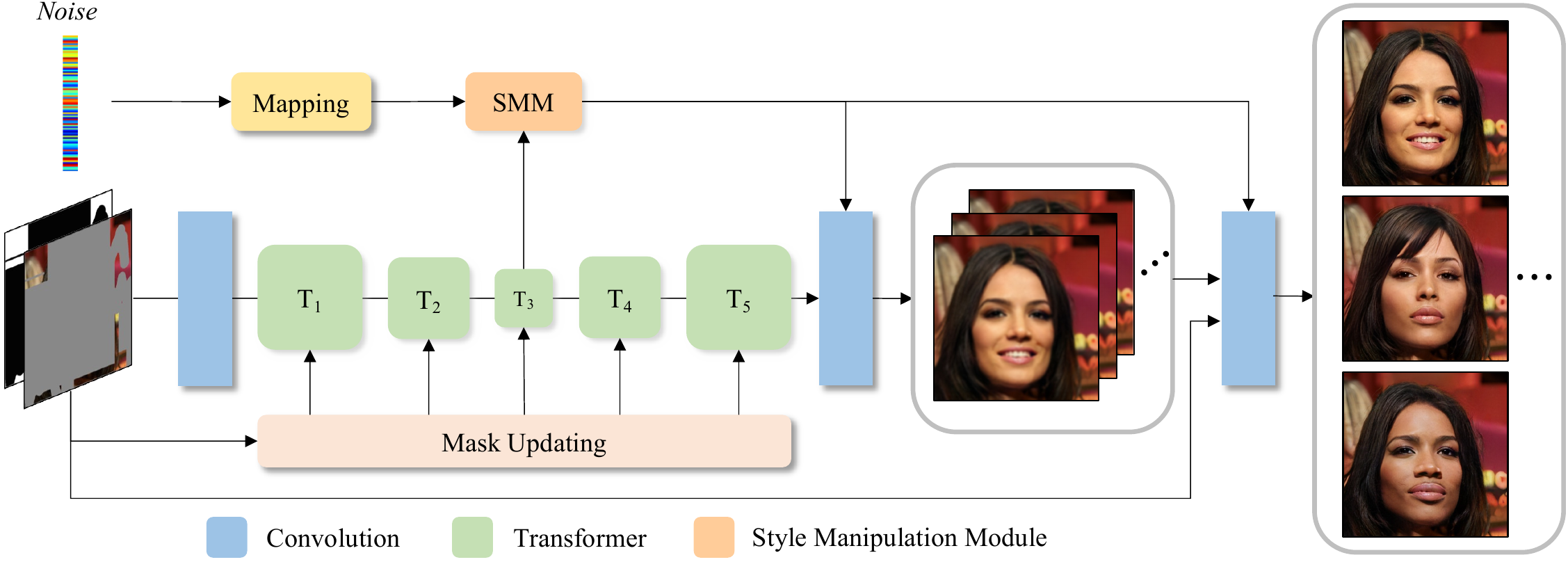}
		\end{center}
		\vspace{-0.1in}
		\caption{The proposed mask-aware transformer (MAT) for pluralistic inpainting, which consists of a convolutional head, a transformer body and a convolutional tail for reconstruction together with a Conv-U-Net for refinement. The mask updating strategy is described in Sec.~\ref{sec:mca}.} 
		\label{fig:framework}
		\vspace{-0.15in}
	\end{figure*}
	
	Recently, researchers switch their focus to more challenging settings, among which the most representative problems are pluralistic generation and large hole filling. For the former, Zheng \etal~\cite{zheng2019pluralistic} propose a probabilistically principled framework with two parallel paths, capable of producing multiple plausible solutions. UCTGAN~\cite{zhao2020uctgan} projects the instance image space and masked image space into a common low-dimensional manifold space via optimizing the KL-divergence to allow diverse generations of missing contents. Later on, ~\cite{wan2021high} and~\cite{yu2021diverse} take advantage of bidirectional attention or auto-regressive transformers to accomplish this goal. Although these methods improve the diversity, their completion and inference performances are limited due to the variational training and raster-scan-order-based generation. On the other hand, some works~\cite{ma2019region,zhao2020large,suvorov2021resolution,zheng2021tfill} are proposed to solve the large hole inpainting problem. For example, CoModGAN~\cite{zhao2020large} leverages the modulation techniques~\cite{chen2018self,karras2019style,karras2020analyzing} to handle large-scale missing regions. In this work, we develop a novel framework to simultaneously achieve high-quality pluralistic generation and large hole filling, bringing the best of long-range context interaction and unconditional generation to the image completion task.
	
	\section{Method}
	\label{sec:method}
	
	Given a masked image, formulated as $\mathbf{I}_{\rm M} = \mathbf{I} \odot \mathbf{M}$, image completion aims to hallucinate visually appealing and semantically appropriate contents for missing areas. In this work, we present a mask-aware transformer (MAT) for large mask inpainting, supporting conditional long-range interactions. 
	Besides, in light of the ill-posed nature of image completion problem, \ie, there could be numerous plausible solutions to fill the large holes, our approach is designed to support pluralistic generation.
	
	\subsection{Overall Architecture}
	\label{sec:arc}
	As shown in Fig.~\ref{fig:framework}, our proposed MAT architecture consists of a convolutional head, a transformer body, a convolutional tail and a style manipulation module, bringing the merits of transformers and convolutions into full play. Specifically, a convolutional head is used to extract tokens, then the main body with five stages of transformer blocks at varying resolutions (with different numbers of tokens) models long-range interactions via the proposed multi-head contextual attention (MCA). For the output tokens from the body, a convolution-based reconstruction module is adopted to upsample the spatial resolution to the input size. Moreover, we adopt another Conv-U-Net to refine high-frequency details, leaning upon the local texture refinement capability and efficiency of CNNs. At last, we design a style manipulation module, enabling the model to deliver diverse predictions by modulating the weights of convolutions. All components in our method are detailed below.

	\subsection{Convolutional Head}
	
	The convolutional head takes in the incompleted image $\mathbf{I}_{\rm M}$ and the given mask $\mathbf{M}$, and produces $\sfrac{1}{8}$ sized feature maps used for tokens. It contains four convolutional layers, one for changing the input dimension and others for downsampling the resolution.
	
	We utilize a convolutional head mainly for two reasons. First, the incorporation of local inductive priors in early visual processing remains vital for better representation~\cite{raghu2021vision} and optimizability~\cite{xiao2021early}. On the other hand, it is designed for fast downsampling to reduce computational complexity and memory cost. Also, we empirically find this design is better than the linear projection head used in ViT~\cite{dosovitskiy2020image}, as validated in the supplementary material. 
	
	\subsection{Transformer Body}
	The transformer body processes tokens by building long-range correspondences. It contains five stages of the proposed adjusted transformer blocks, with an efficient attention mechanism guided by an additional mask.
	
	\subsubsection{Adjusted Transformer Block}
	
\begin{figure}[t]
		\begin{center}
			\includegraphics[width=0.95\linewidth]{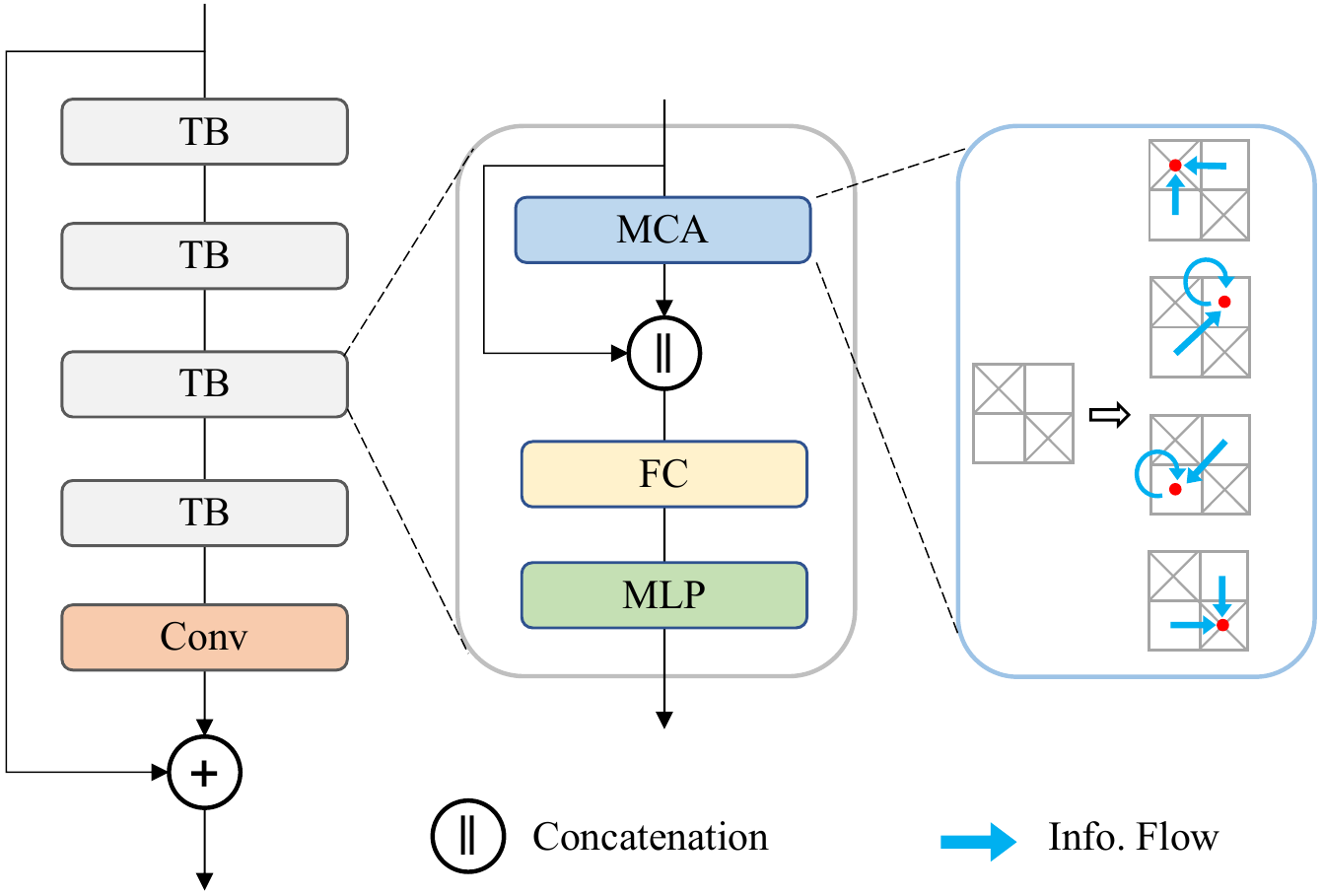}
		\end{center}
		\vspace{-0.1in}
		\caption{Structure of a single transformer stage. ``TB'' refers to an adjusted transformer block and ``MCA'' represents the proposed multi-head contextual attention. The valid tokens are denoted as $\Box$ and invalid tokens are $\boxtimes$. The blue arrow indicates the output of attention is computed as the weighted sum of valid tokens (indicated by blue arrows) while ignoring invalid tokens.}
		\label{fig:tran}
		\vspace{-0.1in}
	\end{figure}

	We propose a new transformer block variant to handle the optimization of masks with large holes. In detail, we remove the layer normalization (LN)~\cite{ba2016layer} and employ fusion learning (using feature concatenation) instead of residual learning. 
	As shown in Fig.~\ref{fig:tran}, we concatenate the input and output of attention and use a fully connected (FC) layer:
	\begin{align}
		\mathbf{X}_{k, \ell}^{'} &= {\rm FC} ( \, \left[ {\rm MCA} (\mathbf{X}_{k, \ell-1}), \ \mathbf{X}_{k, \ell-1} \right] \, ) \,, \\
		\mathbf{X}_{k, \ell} &= {\rm MLP}(\mathbf{X}_{k, \ell}^{'}) \,,
	\end{align}
	where $\mathbf{X}_{k, \ell}$ is the output of the MLP module of the $\ell$-th block in the $k$-th stage. After several transformer blocks, as illustrated in Fig.~\ref{fig:tran}, we adopt a convolution layer with a global residual connection. Note that we abandon the positional embedding in the transformer block since~\cite{xie2021segformer,wu2021cvt} have shown that $3 \times 3$ convolutions are sufficient to provide positional information for transformers. Thus, the flowing only depends on the feature similarity, which promotes long-range interactions.
	
	\vspace{0.08in}
	\noindent{\textbf{Analysis.}}
	The general architecture of transformer~\cite{vaswani2017attention} contains two sub-modules, a multi-head self-attention (MSA) module and an MLP module. Layer normalization is applied before every module and a residual connection~\cite{he2016deep} after every module. Whereas, we observe unstable optimization using the general block when handling large-scale masks, sometimes incurring gradient exploding. We attribute this training issue to the large ratio of invalid tokens (their values are nearly zero). In this circumstance, layer normalization may magnify useless tokens overwhelmingly, leading to unstable training. Besides, residual learning generally encourages the model to learn high-frequency contents. However, considering most tokens are invalid at the beginning, it is difficult to directly learn high-frequency details without proper low-frequency basis in GAN training, which makes the optimization harder. Replacing such residual learning with concatenation leads to obviously superior results, as verified in Sec.~\ref{sec:abl}.

	\subsubsection{Multi-Head Contextual Attention}
	\label{sec:mca}
	
	To handle a large number of tokens (up to 4096 tokens for $512 \times 512$ images) and low fidelity in the given tokens (at most 90$\%$ tokens are useless), our attention module exploits shifted windows~\cite{liu2021Swin} and a dynamical mask, capable of conducting non-local interactions using a few feasible tokens. The output is computed as the weighted sum of valid tokens, as shown in Fig.~\ref{fig:tran}, which is formulated as
	\begin{align}
		{\rm Att}(\mathbf{Q}, \mathbf{K}, \mathbf{V}) = {\rm Softmax}(\frac{\mathbf{Q}\mathbf{K}^{T} + \mathbf{M}^{'}}{\sqrt{d_{k}}})\mathbf{V} \,,
	\end{align}
	where $\mathbf{Q}, \mathbf{K}, \mathbf{V}$ are the query, key, value matrices and $\frac{1}{\sqrt{d_{k}}}$ is the scaling factor. The mask $\mathbf{M}^{'}$ is expressed as:
	\begin{equation}
		\mathbf{M}^{'}_{ij} =\left\{
		\begin{array}{cl}
			0,           & {\rm if \ token} \ j \ \rm{is \ valid} \,, \\
			\text{-}\tau, & {\rm if \ token} \ j \ \rm{is \ invalid} \,,
		\end{array}
		\right.
	\end{equation}
	where $\tau$ is a large positive integer (100 in our experiments). In this case, the aggregation weights of invalid tokens are nearly 0. After each attention, we shift the positions of $w \times w$ sized windows by $(\lfloor \frac{w}{2} \rfloor, \lfloor \frac{w}{2} \rfloor)$ pixels, enabling cross-window connections.
	
\begin{figure}[t]
		\begin{center}
			\includegraphics[width=0.85\linewidth]{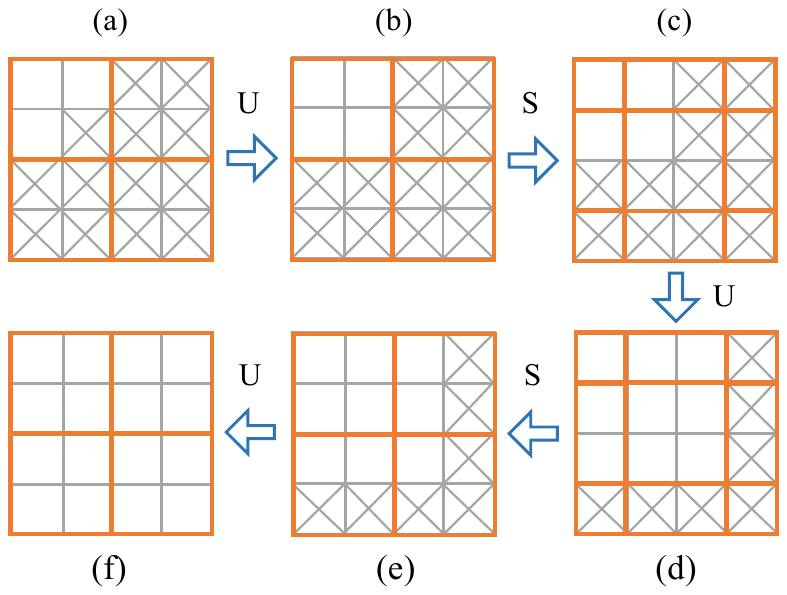}
		\end{center}
		\vspace{-0.2in}
		\caption{Toy example of mask updating. The feature map is initially partitioned into $2 \times 2$ windows (in orange). ``U'' means a mask updating after attention and ``S'' indicates the window shift.  }
		\label{fig:update}
		\vspace{-0.1in}
	\end{figure}

	\vspace{0.08in}
	\noindent{\textbf{Mask Updating Strategy.}}
	The mask ($\mathbf{M}^{'}$) points out whether a token is valid or invalid, which is initialized by the input mask and automatically updated during propagation. The updating follows a rule that all tokens in a window are updated to be valid after attention as long as there is at least one valid token before. If all tokens in a window are invalid, they remain invalid after attention. As shown in Fig.~\ref{fig:update}, going through an attention from (a) to (b), all tokens in the top left window become valid, while tokens in other windows are still invalid. After several times of window shift and attention, the mask is updated to be fully valid.
	
	\vspace{0.08in}
	\noindent{\textbf{Analysis.}}
	For images dominated by missing regions, the default attention strategy not only fails to borrow visible information to inpaint the holes, but also undermines the effective valid pixels. To reduce color discrepancy or blurriness, we propose to only involve valid tokens (selected by a dynamic mask) for computing relations. The effectiveness of our design is manifested in Sec~\ref{sec:abl}.
	
	\subsection{Style Manipulation Module} 
	Inspired by~\cite{chen2018self,karras2019style,karras2020analyzing}, we design a style manipulation module to endow our framework with pluralistic generation. It manipulates the output by changing the weight normalization of convolution layers in the reconstruction procedure with an additional noise input.
	To enhance the representation ability of noise inputs, we enforce the image-conditional style $\mathbf{s}_{c}$ to learn from both the image feature $\mathbf{X}$ and the noise-unconditional style $\mathbf{s}_{u}$, formulated as
	\begin{align}
		& \mathbf{s}_{u} = \mathcal{E}(\mathbf{n}) \,, \\
		& \mathbf{X}^{'} = \mathbf{B} \odot \mathbf{X} + (\textbf{1} - \mathbf{B}) \odot {\rm Resize}(\mathbf{s}_{u}) \,, \\
		& \mathbf{s}_{c} = \mathcal{F}(\mathbf{X}^{'}) \,,
	\end{align}
	where $\mathbf{B}$ is a random binary mask, on which values are set to 1 with a probability of $p$ and to 0 with $1 - p$, $\mathcal{E}$ and $\mathcal{F}$ are mapping functions. As shown in Fig.~\ref{fig:framework}, the style representation is obtained by fusing both style representations:
	\vspace{-0.1in}
	\begin{align}
		\mathbf{s} = \mathcal{A}(\mathbf{s}_{u}, \mathbf{s}_{c}),
	\end{align}
	where $\mathcal{A}$ is a mapping function.
	Then the weights $\mathbf{W}$ of convolutions are baked as
	\begin{align}
		\mathbf{W}^{'}_{ijk} & =  \mathbf{W}_{ijk} \cdot  \mathbf{s}_{i} \,, \\
		\mathbf{W}^{''}_{ijk} & = \mathbf{W}^{'}_{ijk} \bigg/ \sqrt{ \sum\nolimits_{i,k}{\mathbf{W}^{'}_{ijk}}^{2} + \epsilon} \,,
	\end{align}
	where $i,j,k$ denotes the input channels, output channels and spatial footprint of the convolution, respectively, and $\epsilon$ is a small constant. The modulation of different style representations leads to pluralistic outputs. Also, we incorporate the noise injection~\cite{karras2019style} into our framework to further enhance the diversity of generation.
	
	\subsection{Loss Functions}
	\label{sec:lf}
	To improve the quality and diversity of the generation, we adopt the non-saturating adversarial loss~\cite{goodfellow2014generative} for both two stages to optimize our framework, regardless of the pixel-wise MAE or MSE loss that usually leads to averaged blurry results. We also use the $R_{1}$ regularization~\cite{mescheder2018training,ross2018improving}, written as $R_{1}=E_{x} \left \| \nabla D(x)\right \|$. Besides, we adopt the perceptual loss~\cite{johnson2016perceptual} with an empirically low coefficient since we notice it enables easier optimization.
	
	\vspace{0.08in}
	\noindent{\textbf{Adversarial Loss.}}
	We calculate the adversarial loss as
	\begin{align}
		\mathcal{L}_{\rm G} & = - \mathbb{E}_{\hat x} \left[ \log \left(D \left( \hat x \right) \right) \right] \,, \\
		\mathcal{L}_{\rm D} & = - \mathbb{E}_{x} \left[ \log \left( D \left( x \right) \right) \right] - \mathbb{E}_{\hat x} \left[ \log \left( 1 - D \left( \hat x \right) \right) \right] \,,
	\end{align}
	where $x$ and $\hat x$ are the real and generated images. We apply adversarial loss to both two-stage generations in Fig.~\ref{fig:framework}.
	
	\vspace{0.08in}
	\noindent{\textbf{Perceptual Loss.}}
	The perceptual loss is formulated as
	\begin{align}
		\mathcal{L}_{\rm P} = \sum_{i}{ \eta_{i} \left\| \phi_{i} \left( \hat x \right) - \phi_{i} \left ( x \right) \right\|_{1} } \,,
	\end{align}
	where $\phi_{i}(\cdot)$ denotes the layer activation of a pre-trained VGG-19~\cite{Simonyan15} network. We only consider the high-level features of $conv_{4\_4}$ and $conv_{5\_4}$, allowing for variations of inpainted results, with scaling coefficients $\eta_{i}$ as $\frac{1}{4}$ and $\frac{1}{2}$.
	
	\vspace{0.08in}
	\noindent{\textbf{Overall Loss.}}
	The overall loss of the generator is
	\begin{align}
		\mathcal{L} = \mathcal{L}_{{\rm G}} + \gamma R_{1} + \lambda \mathcal{L}_{\rm P} \,.
	\end{align}
	where $\gamma=10$ and $\lambda=0.1$.
	
	\section{Experiments}
	\label{sec:exp}
	
	\subsection{Datasets and Metrics}
	We conduct experiments on the Places365-Standard~\cite{zhou2017places} and the CelebA-HQ~\cite{karras2018progressive} datasets at $512 \times 512$ resolution. Specifically, on the Places dataset, we use the 1.8 million and 36.5 thousand images from train and validation sets to train and evaluate our models, respectively. Images are randomly cropped or padded to $512 \times 512$ size during training while centrally cropped or padded for evaluation. For CelebA-HQ, train and validation splits are organized with 24,183 and 2,993 images. Though trained on $512 \times 512$ images, we show our model generalizes well to a larger resolution in the supplementary material.
		
	\begin{figure*}[t]
		\begin{center}
			\includegraphics[width=1.0\linewidth]{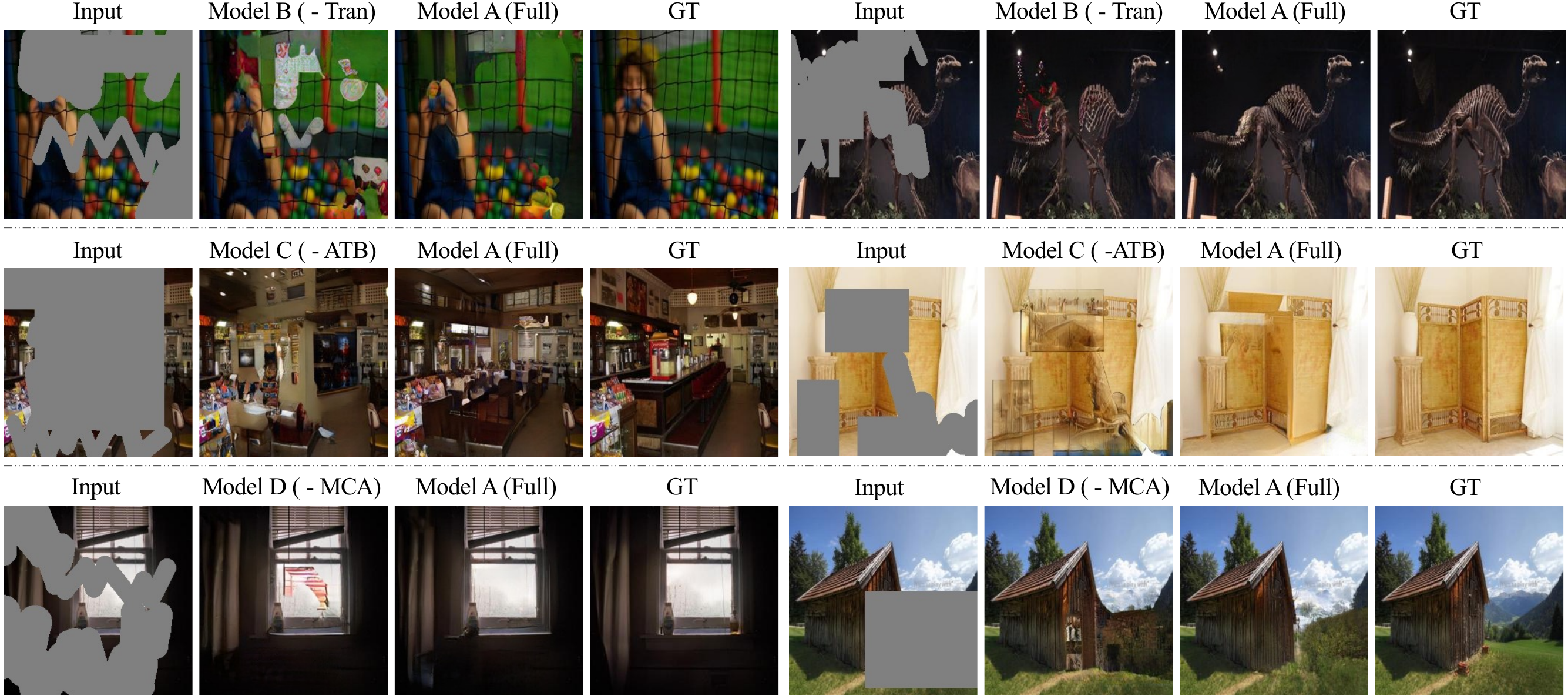}
		\end{center}
		\vspace{-0.2in}
		\caption{Visual examples for ablation study. Model A is our full model, while model B, C, D refer to models replacing transformers with convolutions, using the conventional transformer block and multi-head attention, respectively.}
		\label{fig:ablation}
		\vspace{-0.1in}
	\end{figure*}
	
	In terms of the large hole setting, following~\cite{zhao2020large}, we opt for perceptual metrics including FID~\cite{heusel2017gans}, P-IDS~\cite{zhao2020large} and U-IDS~\cite{zhang2018unreasonable} for evaluation. We suggest that it is inappropriate to use the pixel-wise L1 distance, PSNR and SSIM~\cite{wang2004image}, since preliminary works~\cite{ledig2017photo,sajjadi2017enhancenet} have shown that these metrics correlate weakly with human perception regarding image quality, especially for the ill-posed large-scale image completion problem. Though LPIPS~\cite{zhang2018unreasonable} is calculated in the deep feature space, the pixel-wise evaluation still greatly punishes diverse inpainting systems for large holes. Thus we only use it for reference in the supplementary material. 
	
	
	\subsection{Implementation Details}
	
	In our framework, we set the numbers of convolution channels and FC dimensions to 180 for the head, body, and reconstruction modules. The block numbers and window sizes of 5-level transformer groups are  $\{ 2, 3, 4, 3, 2 \}$ and $\{ 8, 16, 16, 16, 8 \}$, respectively. The last Conv-U-Net firstly downsamples the resolution to $\frac{1}{32}$ and then upsamples to the original size, where the numbers of convolution layers and channels at different scales are borrowed from~\cite{karras2020analyzing}. The mapping network consists of 8 FC layers and the style manipulation module is implemented with convolutions followed with an AvgPool layer. Different from~\cite{wan2021high,zheng2021tfill,yu2021diverse}, our transformer architecture is \textit{without} pre-training.
	
	All experiments are carried out on 8 NVidia V100 GPUs. Following~\cite{zhao2020large}, we train our models for 50M images on Places365-Standard and 25M images on CelebA-HQ. The batch size is 32. We adopt an Adam optimizer with $\beta_{1} = 0$ and $\beta_{2} = 0.99$ and set the learning rate to $1 \times 10^{-3}$. The free-form mask is described in the supplementary file. 

	\subsection{Ablation Study}
	\label{sec:abl}
	In this section, we tease apart which components of our framework contribute most to the final performance. To enable a quick exploration, we only use 100K training images in Places~\cite{zhou2017places} ($\approx 5.6\%$) at $256 \times 256$ resolution and train the models for 5M samples ($10\%$ of the full setting). We adopt the first 10K validation images for evaluation. The quantitative comparison is shown in Table~\ref{tab:ablation}.
	
	
	\vspace{0.05in}
	\noindent{\textbf{Conv-Transformer Architecture.}} We explore whether the long-range context relations modeled by transformers are useful for filling large holes. Replacing the transformer blocks with convolution blocks (model ``B'' in Table~\ref{tab:ablation}), we find an obvious performance drop on all metrics, especially on P-IDS and U-IDS, indicating that the inpainted images lose some fidelity. Moreover, we show some visual examples in Fig.~\ref{fig:ablation}. Compared to the fully convolutional network, our MAT takes advantage of distant context to reconstruct the structure of net and texture of dinosaur skeleton well, showing the effectiveness of long-range interactions.
	
	\begin{table}[t]
		\renewcommand\arraystretch{1.1}
		\begin{center}
			\resizebox{\linewidth}{!}{
				\begin{tabular}{c | l | c | c| c}
					\hline
					Type & Model & FID$\downarrow$ & P-IDS ($\%$)$\uparrow$ & U-IDS($\%$)$\uparrow$ \\
					\hline
					A & Full Model & \textbf{5.97} & \textbf{13.17} & \textbf{29.23}\\
					\hline
					B & - Tran. & 6.21 & 11.30 & 27.39 \\	
					C & - Adjusted Tran. Block & 6.36 & 12.30 & 28.05 \\
					D & - MCA & 6.08 & 13.13 & 29.19 \\
					E & - Style Mani. Module & 6.10 & 11.88 & 27.94 \\
					\hline
					F & - High-Res. Gen. & 6.32 & 12.57 & 28.21 \\
					\hline
				\end{tabular}	
			}
		\end{center}
		\vspace{-0.1in}
		\caption{Ablation study of the framework components. ``A'' represents our full model. ``B'' replaces transformers with convolutions. ``C'' replaces our adjusted transformer block with the original design~\cite{vaswani2017attention}. ``D'' means using the conventional attention strategy.  ``E'' removes the noise style manipulation. ``F'' limits the output size of first-stage generation to $64 \times 64$. }
		\label{tab:ablation}
		\vspace{-0.2in}
	\end{table}

	\begin{figure}[t]
		\begin{center}
			\includegraphics[width=1.0\linewidth]{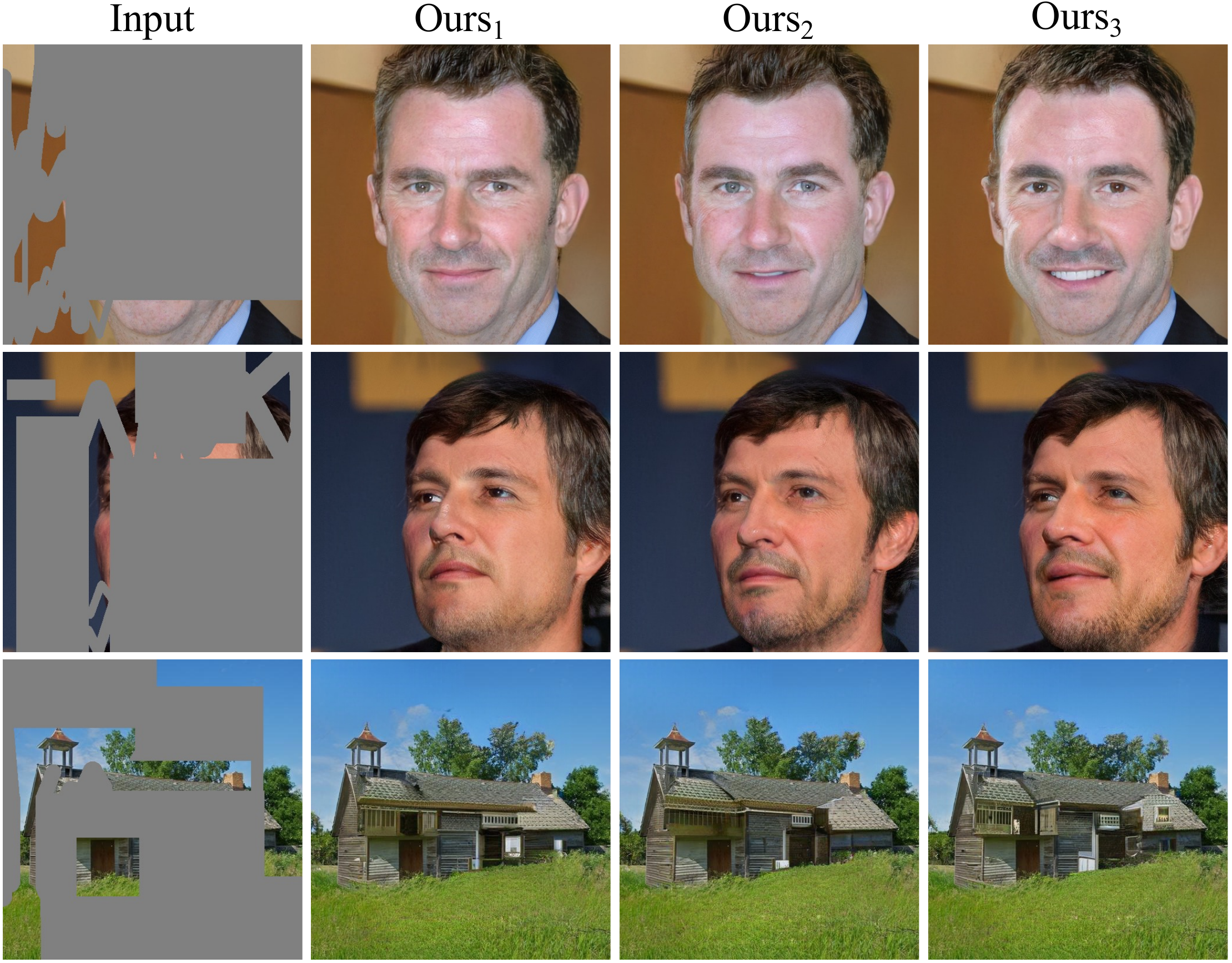}
		\end{center}
		\vspace{-0.1in}
		\caption{Visual examples with different style representations.} 
		\label{fig:interpolation}
		\vspace{-0.1in}
	\end{figure}
	
	\vspace{0.05in}
	\noindent{\textbf{Adjusted Transformer Block.}} In our framework, we develop a novel transformer block since the conventional design easily leads to unstable optimization, in which case we need to lower the learning rate of transformer body. As illustrated in Table~\ref{tab:ablation}, our design (model ``A'') obtains superior performance, 0.39 improvement on FID, than model ``C'' with the original transformer block. As illustrated in Fig.~\ref{fig:ablation}, we notice our design produces more visually appealing results, supporting high-quality image completion. Especially for the first example, even though the missing area is extremely large, our method can still recover a semantically consistent and visually realistic indoor scene.

	\vspace{0.05in}
	\noindent{\textbf{Multi-Head Contextual Attention.}} To quickly fill the missing regions with realistic contents, we propose a multi-head contextual attention (MCA). To make a deeper understanding, we build a model without partial aggregation from valid tokens. The quantitative results are shown as model ``D'' in Table~\ref{tab:ablation}. It is noted that FID drops by 0.1 yet other metrics do not change too much. We suggest the proposed contextual attention is helpful for maintaining color consistency and reducing blurriness. As illustrated in Fig.~\ref{fig:ablation}, the model without MCA generates contents with incorrect colors for the first example, while producing blurry artifacts for the second example. Both the quantitative and qualitative results validate the power of our MCA.
	
	\vspace{0.05in}
	\noindent{\textbf{Style Manipulation Module.}} To deal with large masks, apart from the conditional long-range interaction, we also introduce unconditional generation. To quantify the unconditional generative capability of our framework, we strip the noise style manipulation. From the results of model ``E'' in Table~\ref{tab:ablation}, we find a large gap on P-IDS and U-IDS, showing the modulation of stochastic noise styles further improves the naturalness of completed images.
	
	\vspace{0.05in}
	\noindent{\textbf{High Resolution in Reconstruction.}} Due to quadratically increased computational complexity, existing works~\cite{wan2021high,zheng2021tfill,yu2021diverse} adopt transformers to synthesize low-resolution results, typically $32 \times 32$, for subsequent processing. By contrast, our MAT architecture takes advantage of its computational efficiency to enable high-resolution outputs in the reconstruction stage. As illustrated in Table~\ref{tab:ablation}, our full model ``A'' achieves significant improvement over model ``F'', demonstrating the importance of high-resolution prediction.
	
	
	
	
	\subsection{Comparison with State of the Arts}
	We compare the proposed MAT with a number of state-of-the-art approaches. For a fair comparison, we use publicly available models to test on the same masks. As illustrated in Table~\ref{tab:sota}, MAT achieves state-of-the-art performance on both CelebA-HQ and Places. Especially, even if we only use a subset Places365-Standard (1.8M images) to train our model, much fewer than CoModGAN~\cite{zhao2020large} (8M images) and Big LaMa~\cite{suvorov2021resolution} (4.5M images), MAT still yields promising results. Besides, our method is much more parameter-efficient than the second-best CoModGAN and transformer-based ICT~\cite{wan2021high}. As illustrated in Fig~\ref{fig:qualitative}, compared to other methods, the proposed MAT restores more photo-realistic images with fewer artifacts. For example, our method successfully recovers visually pleasing flowers and regular building structures. 

	\subsection{Pluralistic Generation}
	\label{sec:pluralistic}
	
	The inherent diversity of our framework mainly sources from the style manipulation. As shown in Fig.~\ref{fig:interpolation}, style variants lead to different completions. From the first example in Fig.~\ref{fig:interpolation}, we observe a change from a pursed smile to a toothy laugh. And the second example shows different face contours and appearances. As for the final one, we find different window and roof structures.
	
	
	

	\begin{figure}[t]
		\begin{center}
			\includegraphics[width=1.0\linewidth]{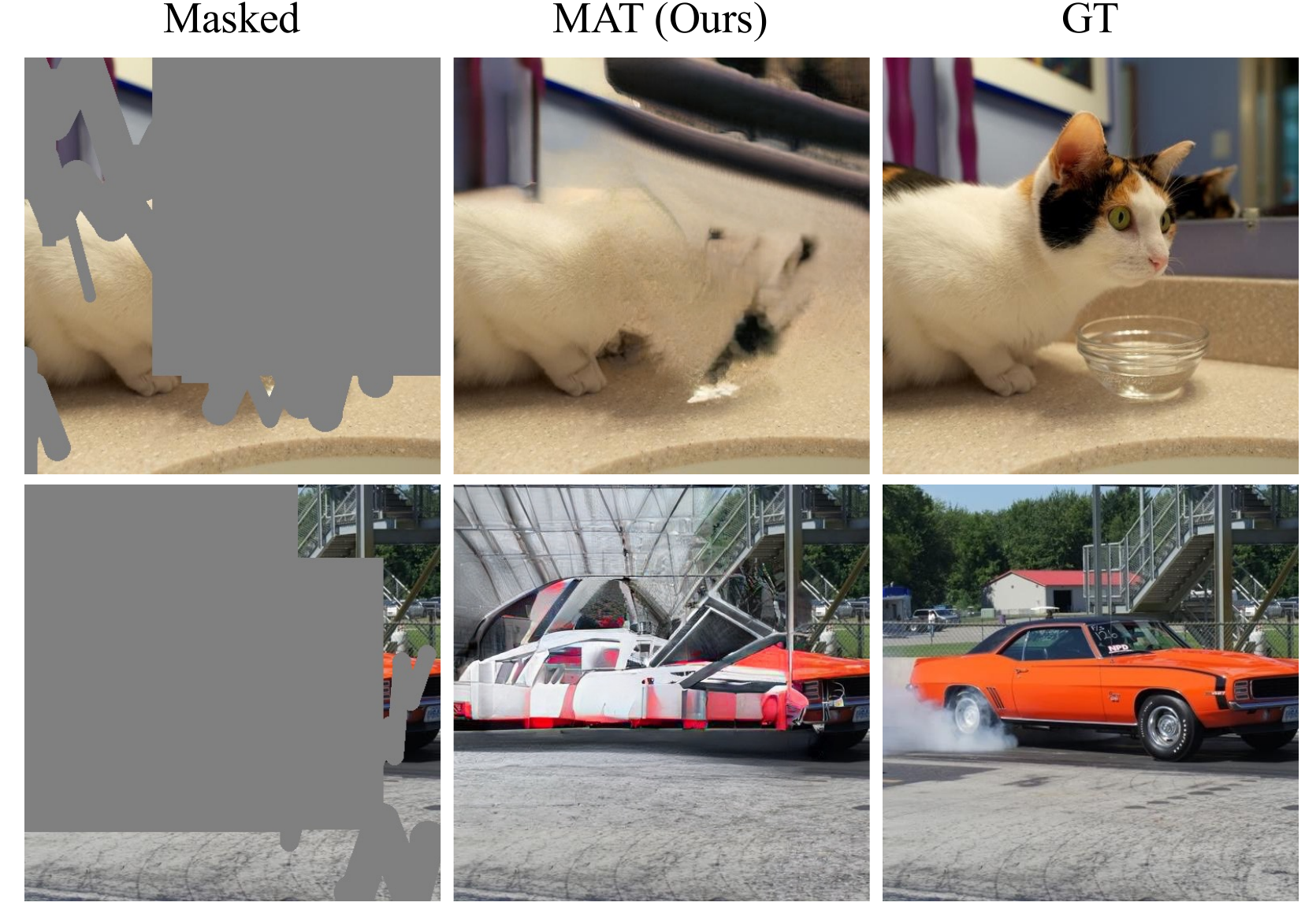}
		\end{center}
		\vspace{-0.1in}
		\caption{Failure cases of our method (MAT). }
		\label{fig:failure}
		\vspace{-0.1in}
	\end{figure}
	
	\begin{table*}[t]
		\renewcommand\arraystretch{1.2}
		\setlength\tabcolsep{2pt}
		\begin{center}
			\resizebox{\linewidth}{!}{
				\begin{tabular}{c | c | c c c | c c c | c c c | c c c }
					\hline
					\multirow{3}{*}{Method} & \multirow{3}{*}{\tabincell{c}{\#Param. \\ $\times 10^6$}} & \multicolumn{6}{|c}{Places ($512 \times 512$)} & \multicolumn{6}{|c}{CelebA-HQ ($512 \times 512$)} \\
					\cline{3-14}
					~ & ~ & \multicolumn{3}{|c|}{Small Mask} & \multicolumn{3}{|c|}{Large Mask} & \multicolumn{3}{|c|}{Small Mask} & \multicolumn{3}{|c}{Large Mask} \\
					\cline{3-14}
					~ & ~ & FID$\downarrow$ & P-IDS($\%$)$\uparrow$ & U-IDS($\%$)$\uparrow$ & FID$\downarrow$& P-IDS($\%$)$\uparrow$ & U-IDS($\%$)$\uparrow$ & FID$\downarrow$ & P-IDS($\%$)$\uparrow$ & U-IDS($\%$)$\uparrow$ & FID$\downarrow$ & P-IDS($\%$)$\uparrow$ & U-IDS($\%$)$\uparrow$ \\
					\hline
					\textbf{MAT (Ours)}$^\dagger$ & \multirow{2}{*}{62} & \textcolor{red}{0.78} & \textcolor{red}{31.72} & \textcolor{red}{43.71} & \textcolor{red}{1.96} & \textcolor{red}{23.42} & \textcolor{red}{38.34} & \multirow{2}{*}{\textcolor{red}{2.86}} & \multirow{2}{*}{\textcolor{red}{21.15}} & \multirow{2}{*}{\textcolor{red}{32.56}} & \multirow{2}{*}{\textcolor{red}{4.86}} & \multirow{2}{*}{\textcolor{red}{13.83}} & \multirow{2}{*}{\textcolor{red}{25.33}} \\
					\textbf{MAT (Ours)} & ~ & 1.07 & \textcolor{blue}{27.42} & \textcolor{blue}{41.93} & \textcolor{blue}{2.90} & 19.03 & 35.36 & ~ & ~ & ~ & ~ & ~ & ~ \\
					\hline
					CoModGAN~\cite{zhao2020large}$^\dagger$ & 109 & 1.10 & 26.95 & 41.88 & 2.92 & \textcolor{blue}{19.64} & \textcolor{blue}{35.78} & \textcolor{blue}{3.26} & \textcolor{blue}{19.65} & \textcolor{blue}{31.41} & \textcolor{blue}{5.65} & \textcolor{blue}{11.23} & \textcolor{blue}{22.54} \\
					LaMa~\cite{suvorov2021resolution}$^\dagger$ & 51/27 & \textcolor{blue}{0.99} & 22.79 & 40.58 & 2.97 & 13.09 & 32.29 & 4.05 & 9.72 & 21.57 & 8.15 & 2.07 & 7.58 \\
					ICT~\cite{wan2021high} & 150 & - & - & - & - & - & - & 6.28 & 2.24 & 9.99 & 12.84 & 0.13 & 0.58 \\
					MADF~\cite{zhu2021image} & 85 & 2.24 & 14.85 & 35.03 & 7.53 & 6.00 & 23.78 & 3.39 & 12.06 & 24.61 & 6.83 & 3.41 & 11.26 \\
					AOT GAN~\cite{zeng2021aggregated} & 15 & 3.19 & 8.07 & 30.94 & 10.64 & 3.07 & 19.92 & 4.65 & 7.92 & 20.45 & 10.82 & 1.94 & 6.97 \\	
					HFill~\cite{yi2020contextual} & 3 & 7.94 & 3.98 & 23.60 & 28.92 & 1.24 & 11.24 & - & - & - & - & - & - \\
					DeepFill v2~\cite{yu2019free} & 4 & 3.02 & 9.17 & 32.56 & 9.27 & 4.01 & 21.32 & 10.11 & 3.11 & 9.52 & 24.42 & 0.17 & 0.42\\
					EdgeConnect~\cite{nazeri2019edgeconnect} & 22 & 4.03 & 5.88 & 27.56 & 12.66 & 1.93 & 15.87 & 10.58 & 4.14 & 12.45 & 39.99 & 0.10 & 0.22 \\
					\hline  
				\end{tabular}
			}
		\end{center}
		\vspace{-0.15in}
		\caption{Quantitative comparison on Places~\cite{zhou2017places} and CelebA-HQ~\cite{karras2018progressive}. ``$\dagger$'': Our Mat, CoModGAN~\cite{zhao2020large} and LaMa~\cite{suvorov2021resolution} use 8M, 8M and 4.5M training images on Places, respectively, while our other model (without ``$\dagger$'') is only trained on a subset (1.8M images). The LaMa models on Places and CelebA are different in size. The results of LPIPS and $256\times 256$ CelebA are provided in the supplementary. The \textcolor{red}{best} and \textcolor{blue}{second best} results are in red and blue.} 
		\label{tab:sota}
	\end{table*}

	\begin{figure*}[t]
		\begin{center}
			\includegraphics[width=1.0\linewidth]{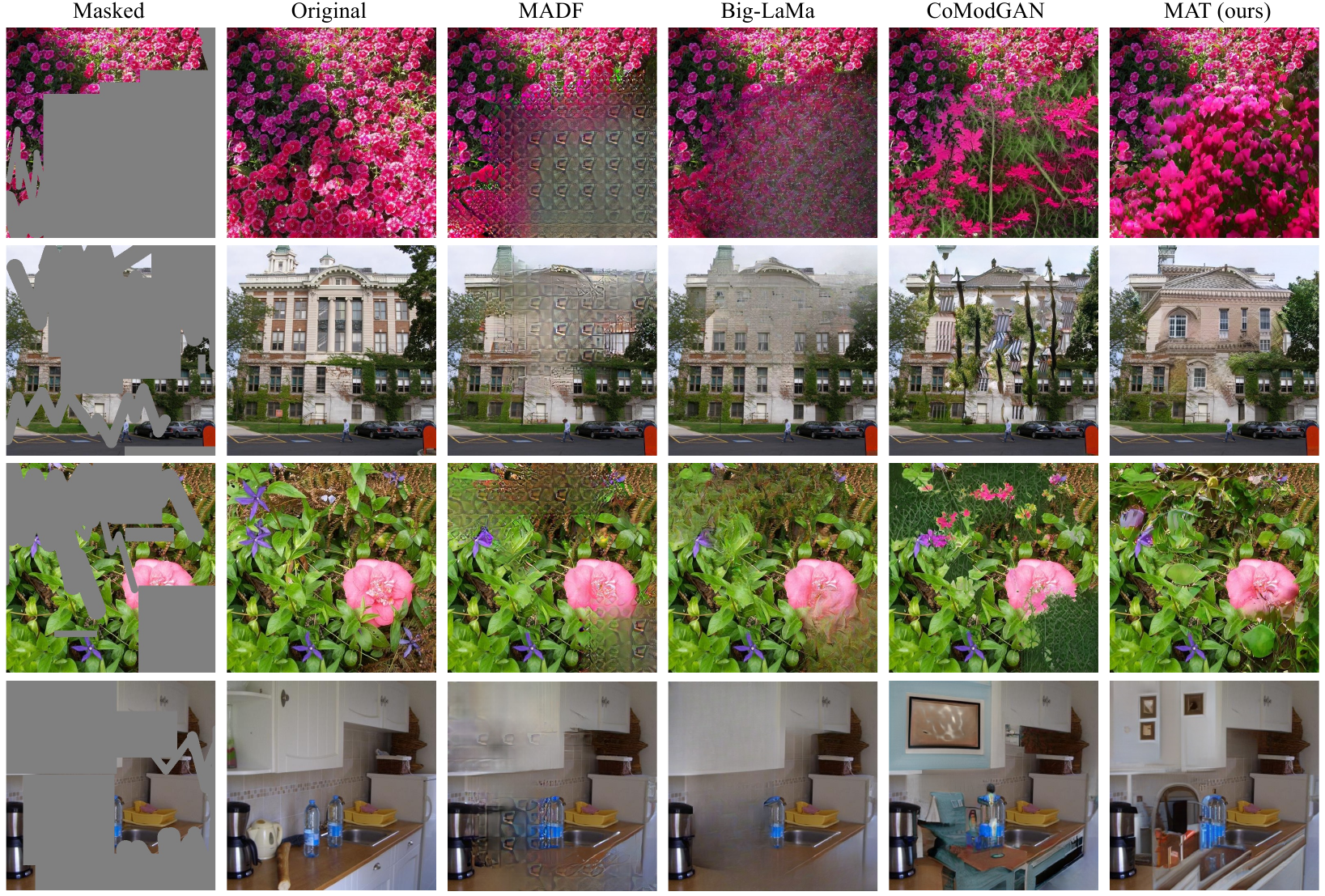}
		\end{center}
		\vspace{-0.15in}
		\caption{Qualitative comparison ($512 \times 512$) with state-of-the-art methods. Our results are more visually realistic, containing more details.}
		\label{fig:qualitative}
		\vspace{-0.05in}
	\end{figure*}
	
	%
	
	\subsection{Limitations and Failure Cases}
	Trained without semantic annotations, MAT usually struggles when processing objects with a variety of shapes, e.g., running animals. As shown in Fig.~\ref{fig:failure}, our method fails to recover the cat and car due to the lack of semantic context understanding. Also, limited by the downsampling and pre-defined window sizes in attention, we need to pad or resize an image to make its size a multiple of 512. 

	\section{Conclusion}
	
	We have presented a mask-aware transformer (MAT) for pluralistic large hole image inpainting. Taking advantage of the proposed adjusted transformer architecture and partial attention mechanism, the proposed MAT achieves state-of-the-art performance on multiple benchmarks. Also, we design a style modulation module to improve the diversity of generation. Extensive qualitative comparisons have demonstrated the superiority of our framework in terms of image quality and diversity. 
	
	{\small
		\bibliographystyle{ieee_fullname}
		\bibliography{egbib}

\begin{thebibliography}{10}\itemsep=-1pt

\bibitem{ba2016layer}
Jimmy~Lei Ba, Jamie~Ryan Kiros, and Geoffrey~E Hinton.
\newblock Layer normalization.
\newblock {\em arXiv preprint arXiv:1607.06450}, 2016.

\bibitem{ballester2001filling}
Coloma Ballester, Marcelo Bertalmio, Vicent Caselles, Guillermo Sapiro, and
  Joan Verdera.
\newblock Filling-in by joint interpolation of vector fields and gray levels.
\newblock {\em TIP}, 10(8):1200--1211, 2001.

\bibitem{barnes2009patchmatch}
Connelly Barnes, Eli Shechtman, Adam Finkelstein, and Dan~B Goldman.
\newblock Patchmatch: A randomized correspondence algorithm for structural
  image editing.
\newblock {\em ToG}, 28(3):24, 2009.

\bibitem{berman2016non}
Dana Berman, Shai Avidan, et~al.
\newblock Non-local image dehazing.
\newblock In {\em CVPR}, pages 1674--1682, 2016.

\bibitem{bertalmio2006strong}
Marcelo Bertalmio.
\newblock Strong-continuation, contrast-invariant inpainting with a third-order
  optimal pde.
\newblock {\em TIP}, 15(7):1934--1938, 2006.

\bibitem{bertalmio2000image}
Marcelo Bertalmio, Guillermo Sapiro, Vincent Caselles, and Coloma Ballester.
\newblock Image inpainting.
\newblock In {\em Proceedings of the 27th annual conference on Computer
  graphics and interactive techniques}, pages 417--424, 2000.

\bibitem{buades2005non}
Antoni Buades, Bartomeu Coll, and J-M Morel.
\newblock A non-local algorithm for image denoising.
\newblock In {\em CVPR}, volume~2, pages 60--65. IEEE, 2005.

\bibitem{chen2018self}
Ting Chen, Mario Lucic, Neil Houlsby, and Sylvain Gelly.
\newblock On self modulation for generative adversarial networks.
\newblock In {\em ICLR}, 2018.

\bibitem{cho2017weakly}
Donghyeon Cho, Jinsun Park, Tae-Hyun Oh, Yu-Wing Tai, and In So~Kweon.
\newblock Weakly-and self-supervised learning for content-aware deep image
  retargeting.
\newblock In {\em ICCV}, pages 4558--4567, 2017.

\bibitem{criminisi2003object}
Antonio Criminisi, Patrick Perez, and Kentaro Toyama.
\newblock Object removal by exemplar-based inpainting.
\newblock In {\em CVPR}, volume~2, pages II--II. IEEE, 2003.

\bibitem{criminisi2004region}
Antonio Criminisi, Patrick P{\'e}rez, and Kentaro Toyama.
\newblock Region filling and object removal by exemplar-based image inpainting.
\newblock {\em TIP}, 13(9):1200--1212, 2004.

\bibitem{ding2018image}
Ding Ding, Sundaresh Ram, and Jeffrey~J Rodr{\'\i}guez.
\newblock Image inpainting using nonlocal texture matching and nonlinear
  filtering.
\newblock {\em TIP}, 28(4):1705--1719, 2018.

\bibitem{dosovitskiy2020image}
Alexey Dosovitskiy, Lucas Beyer, Alexander Kolesnikov, Dirk Weissenborn,
  Xiaohua Zhai, Thomas Unterthiner, Mostafa Dehghani, Matthias Minderer, Georg
  Heigold, Sylvain Gelly, et~al.
\newblock An image is worth 16x16 words: Transformers for image recognition at
  scale.
\newblock In {\em ICLR}, 2020.

\bibitem{fadili2009inpainting}
Mohamed-Jalal Fadili, J-L Starck, and Fionn Murtagh.
\newblock Inpainting and zooming using sparse representations.
\newblock {\em The Computer Journal}, 52(1):64--79, 2009.

\bibitem{ghorai2019multiple}
Mrinmoy Ghorai, Soumitra Samanta, Sekhar Mandal, and Bhabatosh Chanda.
\newblock Multiple pyramids based image inpainting using local patch statistics
  and steering kernel feature.
\newblock {\em TIP}, 28(11):5495--5509, 2019.

\bibitem{goodfellow2014generative}
Ian Goodfellow, Jean Pouget-Abadie, Mehdi Mirza, Bing Xu, David Warde-Farley,
  Sherjil Ozair, Aaron Courville, and Yoshua Bengio.
\newblock Generative adversarial nets.
\newblock {\em NIPS}, 27, 2014.

\bibitem{grossauer2004combined}
Harald Grossauer.
\newblock A combined pde and texture synthesis approach to inpainting.
\newblock In {\em ECCV}, pages 214--224. Springer, 2004.

\bibitem{guo2019progressive}
Zongyu Guo, Zhibo Chen, Tao Yu, Jiale Chen, and Sen Liu.
\newblock Progressive image inpainting with full-resolution residual network.
\newblock In {\em ACMMM}, pages 2496--2504, 2019.

\bibitem{hays2007scene}
James Hays and Alexei~A Efros.
\newblock Scene completion using millions of photographs.
\newblock {\em ToG}, 26(3):4--es, 2007.

\bibitem{he2016deep}
Kaiming He, Xiangyu Zhang, Shaoqing Ren, and Jian Sun.
\newblock Deep residual learning for image recognition.
\newblock In {\em CVPR}, pages 770--778, 2016.

\bibitem{heusel2017gans}
Martin Heusel, Hubert Ramsauer, Thomas Unterthiner, Bernhard Nessler, and Sepp
  Hochreiter.
\newblock Gans trained by a two time-scale update rule converge to a local nash
  equilibrium.
\newblock {\em NIPS}, 30, 2017.

\bibitem{iizuka2017globally}
Satoshi Iizuka, Edgar Simo-Serra, and Hiroshi Ishikawa.
\newblock Globally and locally consistent image completion.
\newblock {\em ToG}, 36(4):1--14, 2017.

\bibitem{jo2019sc}
Youngjoo Jo and Jongyoul Park.
\newblock Sc-fegan: Face editing generative adversarial network with user's
  sketch and color.
\newblock In {\em ICCV}, pages 1745--1753, 2019.

\bibitem{johnson2016perceptual}
Justin Johnson, Alexandre Alahi, and Li Fei-Fei.
\newblock Perceptual losses for real-time style transfer and super-resolution.
\newblock In {\em ECCV}, pages 694--711. Springer, 2016.

\bibitem{karras2018progressive}
Tero Karras, Timo Aila, Samuli Laine, and Jaakko Lehtinen.
\newblock Progressive growing of gans for improved quality, stability, and
  variation.
\newblock In {\em ICLR}, 2018.

\bibitem{karras2019style}
Tero Karras, Samuli Laine, and Timo Aila.
\newblock A style-based generator architecture for generative adversarial
  networks.
\newblock In {\em CVPR}, pages 4401--4410, 2019.

\bibitem{karras2020analyzing}
Tero Karras, Samuli Laine, Miika Aittala, Janne Hellsten, Jaakko Lehtinen, and
  Timo Aila.
\newblock Analyzing and improving the image quality of stylegan.
\newblock In {\em CVPR}, pages 8110--8119, 2020.

\bibitem{le2011examplar}
Olivier Le~Meur, Josselin Gautier, and Christine Guillemot.
\newblock Examplar-based inpainting based on local geometry.
\newblock In {\em ICIP}, pages 3401--3404. IEEE, 2011.

\bibitem{ledig2017photo}
Christian Ledig, Lucas Theis, Ferenc Husz{\'a}r, Jose Caballero, Andrew
  Cunningham, Alejandro Acosta, Andrew Aitken, Alykhan Tejani, Johannes Totz,
  Zehan Wang, et~al.
\newblock Photo-realistic single image super-resolution using a generative
  adversarial network.
\newblock In {\em CVPR}, pages 4681--4690, 2017.

\bibitem{lee2016laplacian}
Joo~Ho Lee, Inchang Choi, and Min~H Kim.
\newblock Laplacian patch-based image synthesis.
\newblock In {\em CVPR}, pages 2727--2735, 2016.

\bibitem{levin2003learning}
Anat Levin, Assaf Zomet, and Yair Weiss.
\newblock Learning how to inpaint from global image statistics.
\newblock In {\em ICCV}, volume~1, pages 305--312, 2003.

\bibitem{li2020recurrent}
Jingyuan Li, Ning Wang, Lefei Zhang, Bo Du, and Dacheng Tao.
\newblock Recurrent feature reasoning for image inpainting.
\newblock In {\em CVPR}, pages 7760--7768, 2020.

\bibitem{liu2018image}
Guilin Liu, Fitsum~A Reda, Kevin~J Shih, Ting-Chun Wang, Andrew Tao, and Bryan
  Catanzaro.
\newblock Image inpainting for irregular holes using partial convolutions.
\newblock In {\em ECCV}, pages 85--100, 2018.

\bibitem{liu2020rethinking}
Hongyu Liu, Bin Jiang, Yibing Song, Wei Huang, and Chao Yang.
\newblock Rethinking image inpainting via a mutual encoder-decoder with feature
  equalizations.
\newblock In {\em ECCV}, pages 725--741. Springer, 2020.

\bibitem{liu2019coherent}
Hongyu Liu, Bin Jiang, Yi Xiao, and Chao Yang.
\newblock Coherent semantic attention for image inpainting.
\newblock In {\em ICCV}, pages 4170--4179, 2019.

\bibitem{liu2021Swin}
Ze Liu, Yutong Lin, Yue Cao, Han Hu, Yixuan Wei, Zheng Zhang, Stephen Lin, and
  Baining Guo.
\newblock Swin transformer: Hierarchical vision transformer using shifted
  windows.
\newblock {\em ICCV}, 2021.

\bibitem{ma2019region}
Yuqing Ma, Xianglong Liu, Shihao Bai, Lei Wang, Aishan Liu, Dacheng Tao, and
  Edwin Hancock.
\newblock Region-wise generative adversarial imageinpainting for large missing
  areas.
\newblock {\em arXiv preprint arXiv:1909.12507}, 2019.

\bibitem{mairal2009non}
Julien Mairal, Francis Bach, Jean Ponce, Guillermo Sapiro, and Andrew
  Zisserman.
\newblock Non-local sparse models for image restoration.
\newblock In {\em ICCV}, pages 2272--2279. IEEE, 2009.

\bibitem{mescheder2018training}
Lars Mescheder, Andreas Geiger, and Sebastian Nowozin.
\newblock Which training methods for gans do actually converge?
\newblock In {\em ICML}, pages 3481--3490. PMLR, 2018.

\bibitem{nazeri2019edgeconnect}
Kamyar Nazeri, Eric Ng, Tony Joseph, Faisal~Z Qureshi, and Mehran Ebrahimi.
\newblock Edgeconnect: Generative image inpainting with adversarial edge
  learning.
\newblock {\em arXiv preprint arXiv:1901.00212}, 2019.

\bibitem{oh2019onion}
Seoung~Wug Oh, Sungho Lee, Joon-Young Lee, and Seon~Joo Kim.
\newblock Onion-peel networks for deep video completion.
\newblock In {\em ICCV}, pages 4403--4412, 2019.

\bibitem{pathak2016context}
Deepak Pathak, Philipp Krahenbuhl, Jeff Donahue, Trevor Darrell, and Alexei~A
  Efros.
\newblock Context encoders: Feature learning by inpainting.
\newblock In {\em CVPR}, pages 2536--2544, 2016.

\bibitem{raghu2021vision}
Maithra Raghu, Thomas Unterthiner, Simon Kornblith, Chiyuan Zhang, and Alexey
  Dosovitskiy.
\newblock Do vision transformers see like convolutional neural networks?
\newblock {\em arXiv preprint arXiv:2108.08810}, 2021.

\bibitem{ren2019structureflow}
Yurui Ren, Xiaoming Yu, Ruonan Zhang, Thomas~H Li, Shan Liu, and Ge Li.
\newblock Structureflow: Image inpainting via structure-aware appearance flow.
\newblock In {\em ICCV}, pages 181--190, 2019.

\bibitem{ronneberger2015u}
Olaf Ronneberger, Philipp Fischer, and Thomas Brox.
\newblock U-net: Convolutional networks for biomedical image segmentation.
\newblock In {\em International Conference on Medical image computing and
  computer-assisted intervention}, pages 234--241. Springer, 2015.

\bibitem{ross2018improving}
Andrew~Slavin Ross and Finale Doshi-Velez.
\newblock Improving the adversarial robustness and interpretability of deep
  neural networks by regularizing their input gradients.
\newblock In {\em AAAI}, 2018.

\bibitem{sajjadi2017enhancenet}
Mehdi~SM Sajjadi, Bernhard Scholkopf, and Michael Hirsch.
\newblock Enhancenet: Single image super-resolution through automated texture
  synthesis.
\newblock In {\em ICCV}, pages 4491--4500, 2017.

\bibitem{Simonyan15}
Karen Simonyan and Andrew Zisserman.
\newblock Very deep convolutional networks for large-scale image recognition.
\newblock In {\em ICLR}, 2015.

\bibitem{song2018spg}
Yuhang Song, Chao Yang, Yeji Shen, Peng Wang, Qin Huang, and C-C~Jay Kuo.
\newblock Spg-net: Segmentation prediction and guidance network for image
  inpainting.
\newblock {\em arXiv preprint arXiv:1805.03356}, 2018.

\bibitem{sun2005image}
Jian Sun, Lu Yuan, Jiaya Jia, and Heung-Yeung Shum.
\newblock Image completion with structure propagation.
\newblock In {\em ACM SIGGRAPH 2005 Papers}, pages 861--868. 2005.

\bibitem{suvorov2021resolution}
Roman Suvorov, Elizaveta Logacheva, Anton Mashikhin, Anastasia Remizova,
  Arsenii Ashukha, Aleksei Silvestrov, Naejin Kong, Harshith Goka, Kiwoong
  Park, and Victor Lempitsky.
\newblock Resolution-robust large mask inpainting with fourier convolutions.
\newblock {\em arXiv preprint arXiv:2109.07161}, 2021.

\bibitem{vaswani2017attention}
Ashish Vaswani, Noam Shazeer, Niki Parmar, Jakob Uszkoreit, Llion Jones,
  Aidan~N Gomez, {\L}ukasz Kaiser, and Illia Polosukhin.
\newblock Attention is all you need.
\newblock In {\em NIPS}, pages 5998--6008, 2017.

\bibitem{wan2020bringing}
Ziyu Wan, Bo Zhang, Dongdong Chen, Pan Zhang, Dong Chen, Jing Liao, and Fang
  Wen.
\newblock Bringing old photos back to life.
\newblock In {\em CVPR}, pages 2747--2757, 2020.

\bibitem{wan2020old}
Ziyu Wan, Bo Zhang, Dongdong Chen, Pan Zhang, Dong Chen, Jing Liao, and Fang
  Wen.
\newblock Old photo restoration via deep latent space translation.
\newblock {\em arXiv preprint arXiv:2009.07047}, 2020.

\bibitem{wan2021high}
Ziyu Wan, Jingbo Zhang, Dongdong Chen, and Jing Liao.
\newblock High-fidelity pluralistic image completion with transformers.
\newblock {\em arXiv preprint arXiv:2103.14031}, 2021.

\bibitem{wang2018non}
Xiaolong Wang, Ross Girshick, Abhinav Gupta, and Kaiming He.
\newblock Non-local neural networks.
\newblock In {\em CVPR}, pages 7794--7803, 2018.

\bibitem{wang2018image}
Yi Wang, Xin Tao, Xiaojuan Qi, Xiaoyong Shen, and Jiaya Jia.
\newblock Image inpainting via generative multi-column convolutional neural
  networks.
\newblock {\em NIPS}, 2018.

\bibitem{wang2004image}
Zhou Wang, Alan~C Bovik, Hamid~R Sheikh, and Eero~P Simoncelli.
\newblock Image quality assessment: from error visibility to structural
  similarity.
\newblock {\em TIP}, 13(4):600--612, 2004.

\bibitem{wu2021cvt}
Haiping Wu, Bin Xiao, Noel Codella, Mengchen Liu, Xiyang Dai, Lu Yuan, and Lei
  Zhang.
\newblock Cvt: Introducing convolutions to vision transformers.
\newblock In {\em ICCV}, pages 22--31, 2021.

\bibitem{xiao2021early}
Tete Xiao, Mannat Singh, Eric Mintun, Trevor Darrell, Piotr Doll{\'a}r, and
  Ross Girshick.
\newblock Early convolutions help transformers see better.
\newblock {\em arXiv preprint arXiv:2106.14881}, 2021.

\bibitem{xie2019image}
Chaohao Xie, Shaohui Liu, Chao Li, Ming-Ming Cheng, Wangmeng Zuo, Xiao Liu,
  Shilei Wen, and Errui Ding.
\newblock Image inpainting with learnable bidirectional attention maps.
\newblock In {\em ICCV}, pages 8858--8867, 2019.

\bibitem{xie2021segformer}
Enze Xie, Wenhai Wang, Zhiding Yu, Anima Anandkumar, Jose~M Alvarez, and Ping
  Luo.
\newblock Segformer: Simple and efficient design for semantic segmentation with
  transformers.
\newblock {\em NeurIPS}, 34, 2021.

\bibitem{xiong2019foreground}
Wei Xiong, Jiahui Yu, Zhe Lin, Jimei Yang, Xin Lu, Connelly Barnes, and Jiebo
  Luo.
\newblock Foreground-aware image inpainting.
\newblock In {\em CVPR}, pages 5840--5848, 2019.

\bibitem{yan2018shift}
Zhaoyi Yan, Xiaoming Li, Mu Li, Wangmeng Zuo, and Shiguang Shan.
\newblock Shift-net: Image inpainting via deep feature rearrangement.
\newblock In {\em ECCV}, pages 1--17, 2018.

\bibitem{yi2020contextual}
Zili Yi, Qiang Tang, Shekoofeh Azizi, Daesik Jang, and Zhan Xu.
\newblock Contextual residual aggregation for ultra high-resolution image
  inpainting.
\newblock In {\em CVPR}, pages 7508--7517, 2020.

\bibitem{yu2018generative}
Jiahui Yu, Zhe Lin, Jimei Yang, Xiaohui Shen, Xin Lu, and Thomas~S Huang.
\newblock Generative image inpainting with contextual attention.
\newblock In {\em CVPR}, pages 5505--5514, 2018.

\bibitem{yu2019free}
Jiahui Yu, Zhe Lin, Jimei Yang, Xiaohui Shen, Xin Lu, and Thomas~S Huang.
\newblock Free-form image inpainting with gated convolution.
\newblock In {\em ICCV}, pages 4471--4480, 2019.

\bibitem{yu2021diverse}
Yingchen Yu, Fangneng Zhan, Rongliang Wu, Jianxiong Pan, Kaiwen Cui, Shijian
  Lu, Feiying Ma, Xuansong Xie, and Chunyan Miao.
\newblock Diverse image inpainting with bidirectional and autoregressive
  transformers.
\newblock {\em arXiv preprint arXiv:2104.12335}, 2021.

\bibitem{zeng2019learning}
Yanhong Zeng, Jianlong Fu, Hongyang Chao, and Baining Guo.
\newblock Learning pyramid-context encoder network for high-quality image
  inpainting.
\newblock In {\em CVPR}, pages 1486--1494, 2019.

\bibitem{zeng2021aggregated}
Yanhong Zeng, Jianlong Fu, Hongyang Chao, and Baining Guo.
\newblock Aggregated contextual transformations for high-resolution image
  inpainting.
\newblock {\em arXiv preprint arXiv:2104.01431}, 2021.

\bibitem{zeng2020high}
Yu Zeng, Zhe Lin, Jimei Yang, Jianming Zhang, Eli Shechtman, and Huchuan Lu.
\newblock High-resolution image inpainting with iterative confidence feedback
  and guided upsampling.
\newblock In {\em ECCV}, pages 1--17. Springer, 2020.

\bibitem{zhang2018semantic}
Haoran Zhang, Zhenzhen Hu, Changzhi Luo, Wangmeng Zuo, and Meng Wang.
\newblock Semantic image inpainting with progressive generative networks.
\newblock In {\em ACMMM}, pages 1939--1947, 2018.

\bibitem{zhang2018unreasonable}
Richard Zhang, Phillip Isola, Alexei~A Efros, Eli Shechtman, and Oliver Wang.
\newblock The unreasonable effectiveness of deep features as a perceptual
  metric.
\newblock In {\em CVPR}, pages 586--595, 2018.

\bibitem{zhao2020uctgan}
Lei Zhao, Qihang Mo, Sihuan Lin, Zhizhong Wang, Zhiwen Zuo, Haibo Chen, Wei
  Xing, and Dongming Lu.
\newblock Uctgan: Diverse image inpainting based on unsupervised cross-space
  translation.
\newblock In {\em CVPR}, pages 5741--5750, 2020.

\bibitem{zhao2020large}
Shengyu Zhao, Jonathan Cui, Yilun Sheng, Yue Dong, Xiao Liang, I Eric, Chao
  Chang, and Yan Xu.
\newblock Large scale image completion via co-modulated generative adversarial
  networks.
\newblock In {\em ICLR}, 2020.

\bibitem{zheng2019pluralistic}
Chuanxia Zheng, Tat-Jen Cham, and Jianfei Cai.
\newblock Pluralistic image completion.
\newblock In {\em CVPR}, pages 1438--1447, 2019.

\bibitem{zheng2021tfill}
Chuanxia Zheng, Tat-Jen Cham, and Jianfei Cai.
\newblock Tfill: Image completion via a transformer-based architecture.
\newblock {\em arXiv preprint arXiv:2104.00845}, 2021.

\bibitem{zhou2017places}
Bolei Zhou, Agata Lapedriza, Aditya Khosla, Aude Oliva, and Antonio Torralba.
\newblock Places: A 10 million image database for scene recognition.
\newblock {\em PAMI}, 40(6):1452--1464, 2017.

\bibitem{zhu2021image}
Manyu Zhu, Dongliang He, Xin Li, Chao Li, Fu Li, Xiao Liu, Errui Ding, and
  Zhaoxiang Zhang.
\newblock Image inpainting by end-to-end cascaded refinement with mask
  awareness.
\newblock {\em TIP}, 30:4855--4866, 2021.

\end{thebibliography}
	}
	
	\clearpage
	
	\renewcommand\thesection{\Alph{section}}
	\renewcommand\thesubsection{\thesection.\arabic{subsection}}
	\renewcommand\thefigure{\Alph{section}.\arabic{figure}}
	\renewcommand\thetable{\Alph{section}.\arabic{table}} 
	
	\setcounter{section}{0}
	\setcounter{figure}{0}
	\setcounter{table}{0}
	
	\twocolumn[
	\begin{@twocolumnfalse}
		\begin{center}
			\noindent{\Large{\textbf{MAT: Mask-Aware Transformer for Large Hole Image Inpainting \\ 
						\vspace{0.1in}
						(Supplementary Material)}}}
		\end{center}
		\vspace{0.3in}
	\end{@twocolumnfalse}
	]

	\section{Network Architecture}
	\label{sec:net}
	As illustrated in Sec.~\textcolor{red}{3.1}, the proposed MAT is a two-stage framework, where the first stage consists of a convolutional head, a transformer body and a convolutional reconstruction tail while the second stage is a Conv-U-Net. And the discriminator follows the design of CoModGAN~\cite{zhao2020large}.
	
	Given an $H \times W$ input, the head first applies a convolution to change the number of channels from 4 (image 3 + mask 1) to 180 and then adopts three strided convolutions (stride = 2) to downsample the feature size to $\frac{H}{8} \times \frac{W}{8}$. The feature is transformed to tokens as input to the transformer body. The body is composed of five stages of transformer blocks, where the block numbers are $\{2, 3, 4, 3, 2\}$ and the corresponding feature sizes are $\{\frac{H}{8} \times \frac{W}{8}, \frac{H}{16} \times \frac{W}{16}, \frac{H}{32} \times \frac{W}{32}, \frac{H}{16} \times \frac{W}{16}, \frac{H}{8} \times \frac{W}{8} \}$. The downsampling and upsampling are realized by convolutions. The detailed structure of a transformer block is shown in Sec.~\textcolor{red}{3.3}. Then the output tokens from the body are converted to a 2D feature, passed to the reconstruction tail. The convolutional tail upsamples the feature size from $\frac{H}{8} \times \frac{W}{8}$ to $H \times W$ and generates a completed image, during which style modulation is applied to all layers to enable pluralistic generation.
	
	The second-stage Conv-U-Net takes in the coarse prediction and the input mask for subsequent high-fidelity detail rendering. It first downsamples the feature size to $\frac{H}{32} \times \frac{W}{32}$ and then upsamples the size back to $H \times W$. Shortcut connections are adopted at each resolution. The number of convolution channels in the encoder starts from 64 and is doubled after each downsampling, with a maximum of 512, while the decoder uses a symmetrical setting. Besides, all decoding layers are modulated by the image-conditional and noise-unconditional style representations.
	
	\begin{figure}[t]
		\begin{center}
			\includegraphics[width=1.0\linewidth]{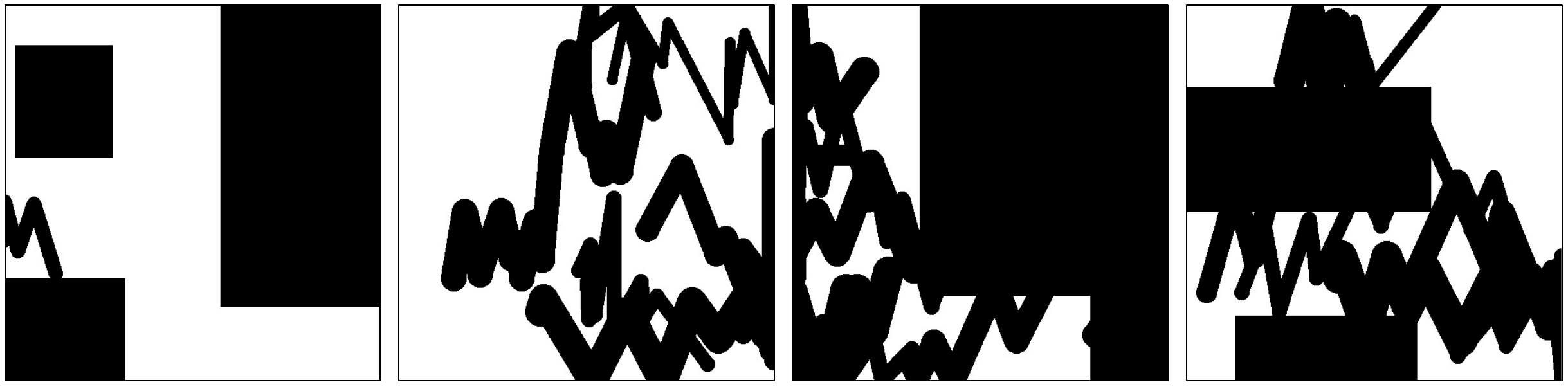}
		\end{center}
		\vspace{-0.1in}
		\caption{Examples of free-form masks ($512 \times 512$). Visible and invisible pixels are in white and black colors.}
		\label{fig:mask}
	\end{figure}
	
	\begin{figure}[t]
		\begin{center}
			\includegraphics[width=1.0\linewidth]{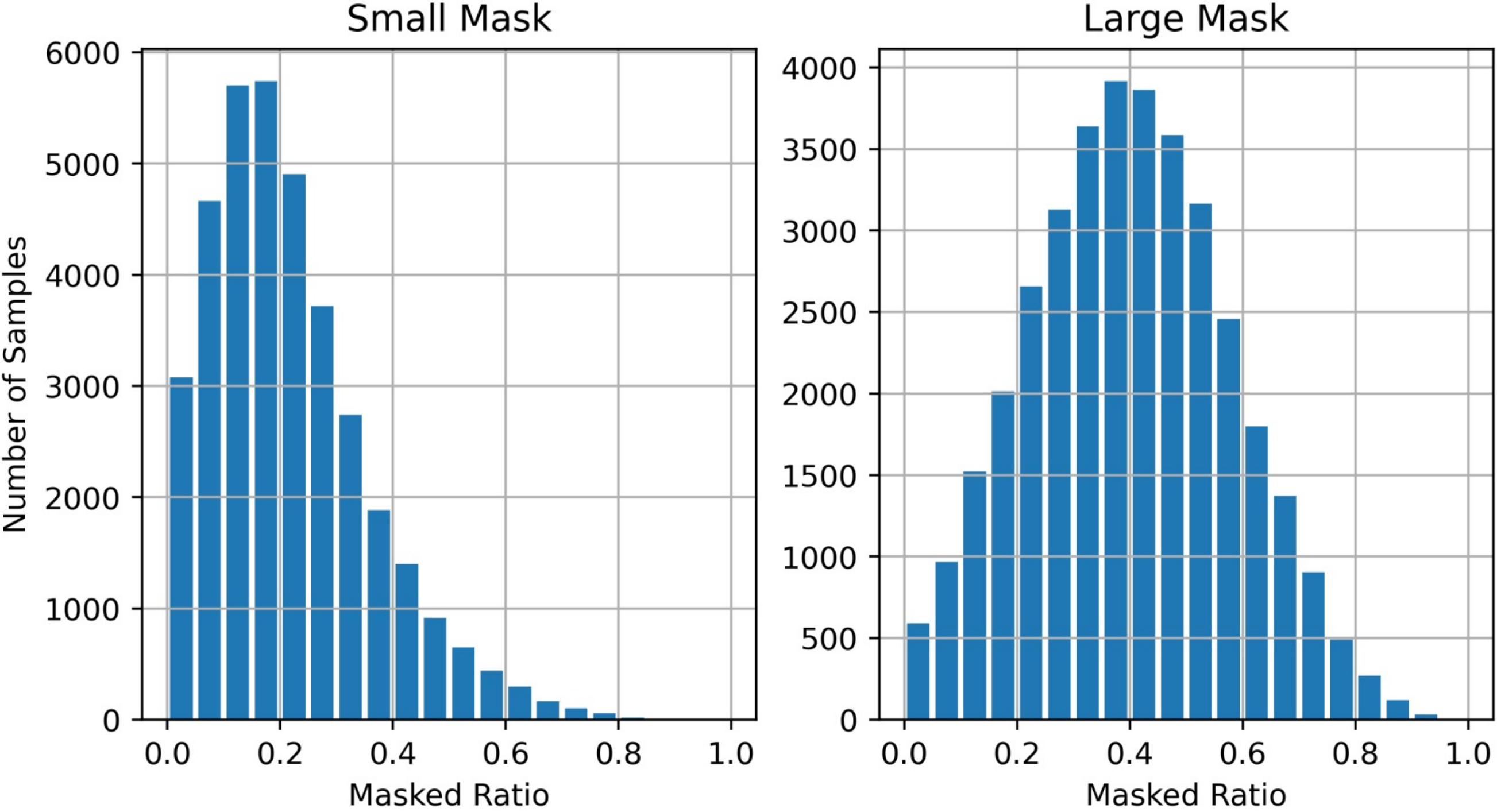}
		\end{center}
		\vspace{-0.15in}
		\caption{Small and large mask ($512 \times 512$) statistics on the Places Val set~\cite{zhou2017places}. The are totally 36500 masks.}
		\label{fig:statistics}
		\vspace{-0.2in}
	\end{figure}
	
	\section{Free-Form Mask Sampling and Statistics}
	\label{sec:mask}
	
	Referring to DeepFill v2~\cite{yu2019free}, we sample rectangles and brush strokes with random sizes, shapes and locations to generate free-form masks. During training, we use a large mask sampling strategy. The number of up to full-size or half-size rectangles is uniformly sampled within $\left[0, 3\right]$ or $\left[0, 5\right]$. The number of strokes is randomly sampled within [0, 9], with a random brush width within $\left[12, 48\right]$ and vertex number within $\left[4, 18\right]$. During testing, apart from the large mask setup, we also introduce a small mask sampling strategy, where the number of up to full-size or half-size rectangles is within $\left[0, 2\right]$ or $\left[0, 3\right]$ and the number of strokes is within [0, 4], while other settings remain unchanged. Note that our model is trained on large masks and is evaluated on both small and large mask settings. As shown in Fig.~\ref{fig:statistics}, we present the mask statistics on the Places Val set~\cite{zhou2017places} that is used for evaluation. It is observed that large masks are very aggressive and diverse.
	
	\section{Tokenization}
	\label{sec:token}
	
\begin{figure}[t]
		\begin{center}
			\includegraphics[width=1.0\linewidth]{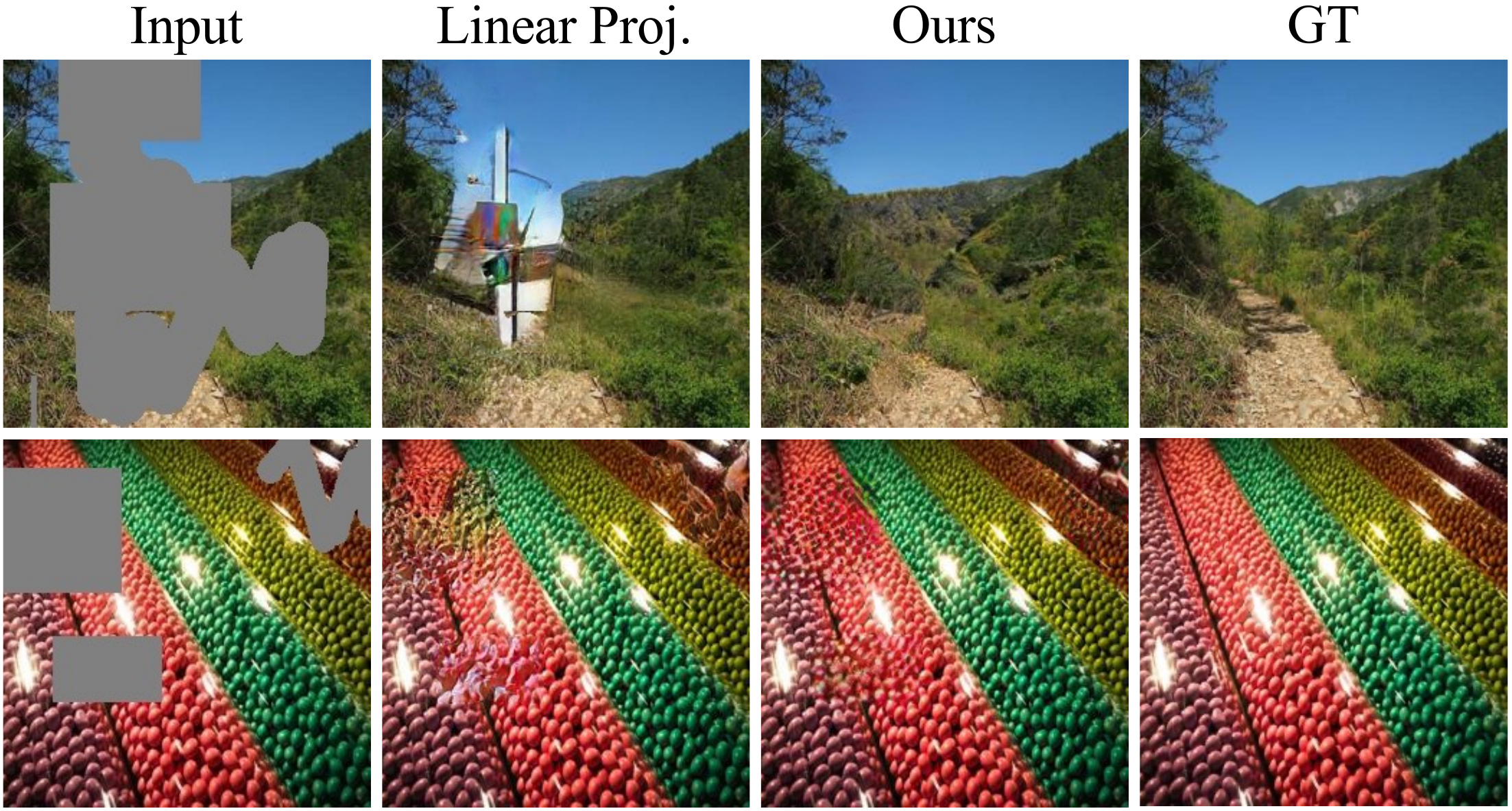}
		\end{center}
		\vspace{-0.15in}
		\caption{Qualitative comparison between linear projection and stacked convolutions (ours) for tokenization.}
		\label{fig:token}
		\vspace{-0.1in}
	\end{figure}

	As described in Sec.~\ref{sec:net}, we adopt a stack of convolutions (the convolutional head) to extract tokens for the transformer body, which is specially tailored to the inpainting problem. Compared to the linear projection of ViT~\cite{dosovitskiy2020image}, our design owns two merits. First, stacked convolutions can gradually fill the holes, producing more effective tokens. Second, the multi-scale downsampled features can be passed to the decoder through shortcut connections, improving the optimization. As illustrated in Table~\ref{tab:lp} and Fig.~\ref{fig:token}, stacked convolutions obtain obviously superior results. The model using linear projection is more likely to generate unpleasing artifacts and fail to borrow surrounding textures to fill the holes, while our MAT successfully recovers high-fidelity contents. Both the quantitative and qualitative results demonstrate the effectiveness of our MAT.
	
	\begin{table}[t]
		\renewcommand\arraystretch{1.1}
		\small
		\begin{center}
			\begin{tabular}{ l | c | c| c}
				\hline
				Model & FID$\downarrow$ & P-IDS ($\%$)$\uparrow$ & U-IDS($\%$)$\uparrow$ \\
				\hline
				Stacked Conv. (Ours) & \textbf{5.97} & \textbf{13.17} & \textbf{29.23}\\
				Linear Projection & 10.54 & 5.77 & 20.86 \\
				\hline
			\end{tabular}	
		\end{center}
		\vspace{-0.15in}
		\caption{Quantitative comparison between linear projection and stacked convolutions for token extraction. We use the same training setting as the ablation study (Sec.~\textcolor{red}{4.3}).}
		\label{tab:lp}
	\end{table}
	
	\section{Model Configuration}
	Following the same experimental setting as ablation study, we explore several model variants in terms of feature width, block number and window size of the transformer body, leaving Conv-U-Net unchanged. The results are shown in the Table~\ref{tab:config}. The performance is positively correlated to the model capacity and attention range. 
	
	\begin{table}[t]
		\small
		\centering
		\resizebox{\linewidth}{!}{
			\begin{tabular}{c | c | c | c | c }
				\hline
				Model & Feature Dim. & Block Num. & Window Size & FID$\downarrow$ \\
				\hline
				Ours & 180 & $\{2, 3, 4, 3, 2\}$ & $\{{8, 16, 16, 16, 8}\}$ & \textbf{5.97} \\
				\hline
				V1 & 90 & $\{{2, 3, 4, 3, 2}\}$ & $\{{8, 16, 16, 16, 8}\}$ & 6.28 \\
				V2 & 180 & $\{{1, 1, 2, 1, 1}\}$ & $\{{8, 16, 16, 16, 8}\}$ & 6.18 \\
				V3 & 180 & $\{{2, 3, 4, 3, 2}\}$ & $\{{8, 8, 8, 8, 8}\}$ & 6.09 \\
				\hline
			\end{tabular}
		}
		\caption{Ablation study on model configuration.}
		\label{tab:config}
	\end{table}
	
	\section{CelebA-HQ 256 $\times$ 256 Results}
	
	We provide the quantitative results on $256 \times 256$ CelebA-HQ~\cite{karras2018progressive}. As illustrated in Table~\ref{tab:celeba}, our MAT yields significant improvements on FID~\cite{heusel2017gans}, P-IDS~\cite{zhao2020large} and U-IDS~\cite{zhang2018unreasonable} metrics over other methods.
	
	\section{LPIPS Results}
	As discussed in Sec.~\textcolor{red}{4.1}, LPIPS~\cite{zhang2018unreasonable} is not an appropriate measure for large mask inpainting, especially for pluralistic generation systems, since there could be numerous plausible solutions to fill the holes. Therefore, we provide the LPIPS results only for reference. As shown in Table~\ref{tab:lpips}, our method achieves superior or comparable performance on the CelebA-HQ~\cite{karras2018progressive} and Places~\cite{zhou2017places} datasets. \textit{Note that we only use 22.5\% of full data to train our Places model}.
	
	\begin{table}[t]
		\renewcommand\arraystretch{1.2}
		\setlength\tabcolsep{3pt}
		\begin{center}
			\resizebox{\linewidth}{!}{
				\begin{tabular}{c | c c c | c c c}
					\hline
					\multirow{2}{*}{Method} & \multicolumn{3}{|c}{Small Mask} & \multicolumn{3}{|c}{Large Mask} \\
					\cline{2-7}
					~ & FID$\downarrow$ & P-IDS$\uparrow$ & U-IDS$\uparrow$ & FID$\downarrow$ & P-IDS$\uparrow$ & U-IDS$\uparrow$ \\
					\hline
					MAT (Ours) & \textbf{2.94} & \textbf{20.88} & \textbf{32.01} & \textbf{5.16} & \textbf{13.90} & \textbf{25.13} \\
					\hline
					LaMa~\cite{suvorov2021resolution} & 3.98 & 8.82 & 22.57  & 8.75 & 2.34 & 8.77 \\
					ICT~\cite{wan2021high} & 5.24 & 4.51 & 17.39 & 10.92 & 0.90 & 5.23 \\
					MADF~\cite{zhu2021image} & 10.43 & 6.25 & 14.62 & 23.59 & 0.50 & 1.44 \\
					AOT GAN~\cite{zeng2021aggregated} & 9.64 & 5.61 & 14.62 & 22.91 & 0.47 & 1.65 \\
					DeepFill v2~\cite{yu2019free} & 5.69 & 6.62 & 16.82 & 13.23 & 0.84 & 2.62 \\
					EdgeConnect~\cite{nazeri2019edgeconnect} & 5.24 & 5.61 & 15.65 & 12.16 & 0.84 & 2.31 \\
					\hline  
				\end{tabular}
			}
		\end{center}
		\vspace{-0.1in}
		\caption{Quantitative results on CelebA-HQ at $256 \times 256$ size. The results of P-IDS and U-IDS are shown in percentage ($\%$).}
		\label{tab:celeba}
	\end{table}
	
	\begin{table}[t]
		\renewcommand\arraystretch{1.1}
		\begin{center}
			\resizebox{\linewidth}{!}{
				\begin{tabular}{c | c | c c | c c }
					\hline
					\multirow{2}{*}{Method} & \#Param. & \multicolumn{2}{|c}{CelebA-HQ} & \multicolumn{2}{|c}{Places} \\
					\cline{3-6}
					~ & $\times 10^6$ & Small & Large & Small & Large \\
					\hline
					MAT (Ours) & 60 & \textbf{0.065} & \textbf{0.125} & 0.099 & 0.189 \\
					\hline
					CoModGAN~\cite{zhao2020large}$^\dagger$ & 109 & 0.073 & 0.140 & 0.101 & 0.192 \\
					LaMa~\cite{suvorov2021resolution}$^\dagger$ & 27/51 & 0.075 & 0.143 & \textbf{0.086} & \textbf{0.166} \\
					ICT~\cite{wan2021high} & 150 & 0.105 & 0.195 & - & - \\
					MADF~\cite{zhu2021image} & 85 & 0.068 & 0.130 & 0.095 & 0.181\\
					AOT GAN~\cite{zeng2021aggregated} & 15 & 0.074 & 0.145 & 0.101 & 0.195 \\
					HFill~\cite{yi2020contextual} & 3 & - & - & 0.148 & 0.284 \\
					DeepFill v2~\cite{yu2019free} & 4 & 0.117 & 0.221 & 0.113 & 0.213 \\
					EdgeConnect~\cite{nazeri2019edgeconnect} & 22 & 0.101 & 0.208 & 0.114 & 0.275 \\
					\hline  
				\end{tabular}
			}
		\end{center}
		\vspace{-0.1in}
		\caption{LPIPS~\cite{zhang2018unreasonable} comparison on $512 \times 512$ CelebA-HQ~\cite{karras2018progressive} and Places~\cite{zhou2017places} datasets. ``$\dagger$'': CoModGAN~\cite{zhao2020large} and LaMa~\cite{suvorov2021resolution} use 8M and 4.5M Places images to train their models, while our model is only trained on Places365-Standard (1.8M images). The LaMa models on CelebA-HQ and Places are different in size. }
		\label{tab:lpips}
	\end{table}
	
	\section{Generalization to A Higher Resolution}
	Though trained on $512 \times 512$ images, our model generalizes well to larger resolutions. For example, we transfer our model and Big LaMa~\cite{suvorov2021resolution} trained at $512 \times 512$ resolution to $1024 \times 1024$. Compared to Big LaMa (FID$\downarrow$ 6.31, PIDS$\uparrow$ 4.98$\%$), our model (FID$\downarrow$ 5.83, P-IDS$\uparrow$ 9.51$\%$) obtains superior results on Places under the large mask setting. We suggest that maintaining a resolution consistency during training and testing yields better visual quality.
	
	\section{Diversity-Fidelity Tradeoff}
	To evaluate the fidelity and diversity, apart from FID (depending on both diversity and fidelity), we also follow~\cite{kynkaanniemi2019improved,dhariwal2021diffusion} to use Improved Precision and Recall to separately measure sample fidelity (precision) and diversity (recall). As shown in Table~\ref{tab:tradeoff}, our method obtains better FID, higher recall yet slightly lower precision compared to CoModGAN on Places. It is noted that we use much less training data.
	
	\begin{table}[t]
		\small
		\centering
		\resizebox{\linewidth}{!}{
			\begin{tabular}{c | c | c | c | c}
				\hline
				Method & Training Data & FID$\downarrow$ & Precision$\uparrow$ & Recall$\uparrow$ \\
				\hline
				\textbf{MAT (Ours)} & \textbf{1.8M} & \textbf{2.90} & 0.925 & \textbf{0.951} \\
				CoModGAN & 8M & 2.92 & \textbf{0.929} & 0.942 \\
				\hline
			\end{tabular}
		}
		\caption{Precision and Recall results of our MAT and CoModGAN on Places.}
		\label{tab:tradeoff}
	\end{table}
	
	
	%
	
	\section{Additional Qualitative Results}
	\label{sec:visual}
	
\begin{figure*}[t]
		\begin{center}
			\includegraphics[width=1.0\linewidth]{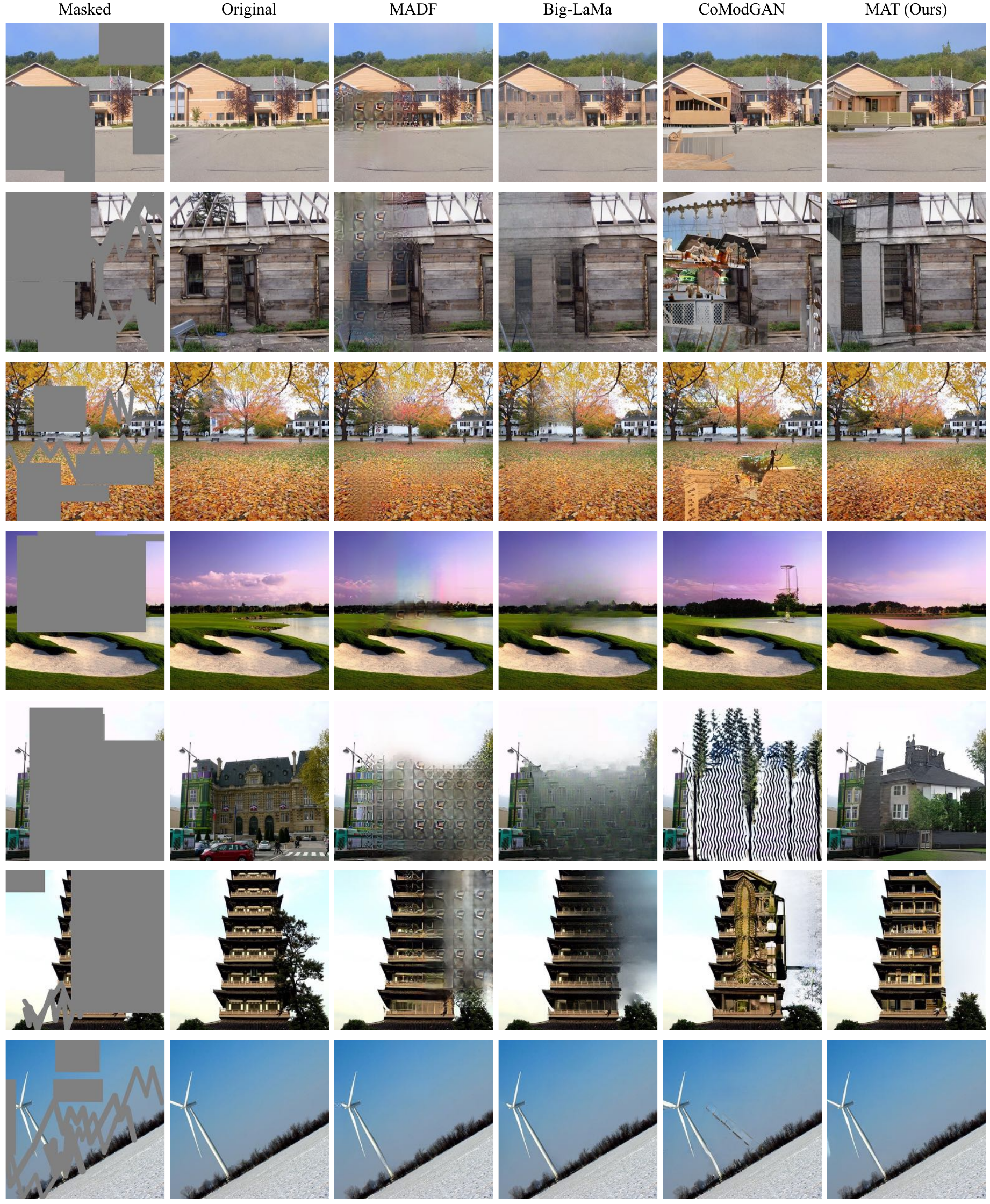}
		\end{center}
		\caption{Qualitative comparison ($512 \times 512$) with state-of-the-art methods on the Places dataset. Zoom in for a better view.}
		\label{fig:sota1}
	\end{figure*}

\begin{figure*}[t]
		\begin{center}
			\includegraphics[width=1.0\linewidth]{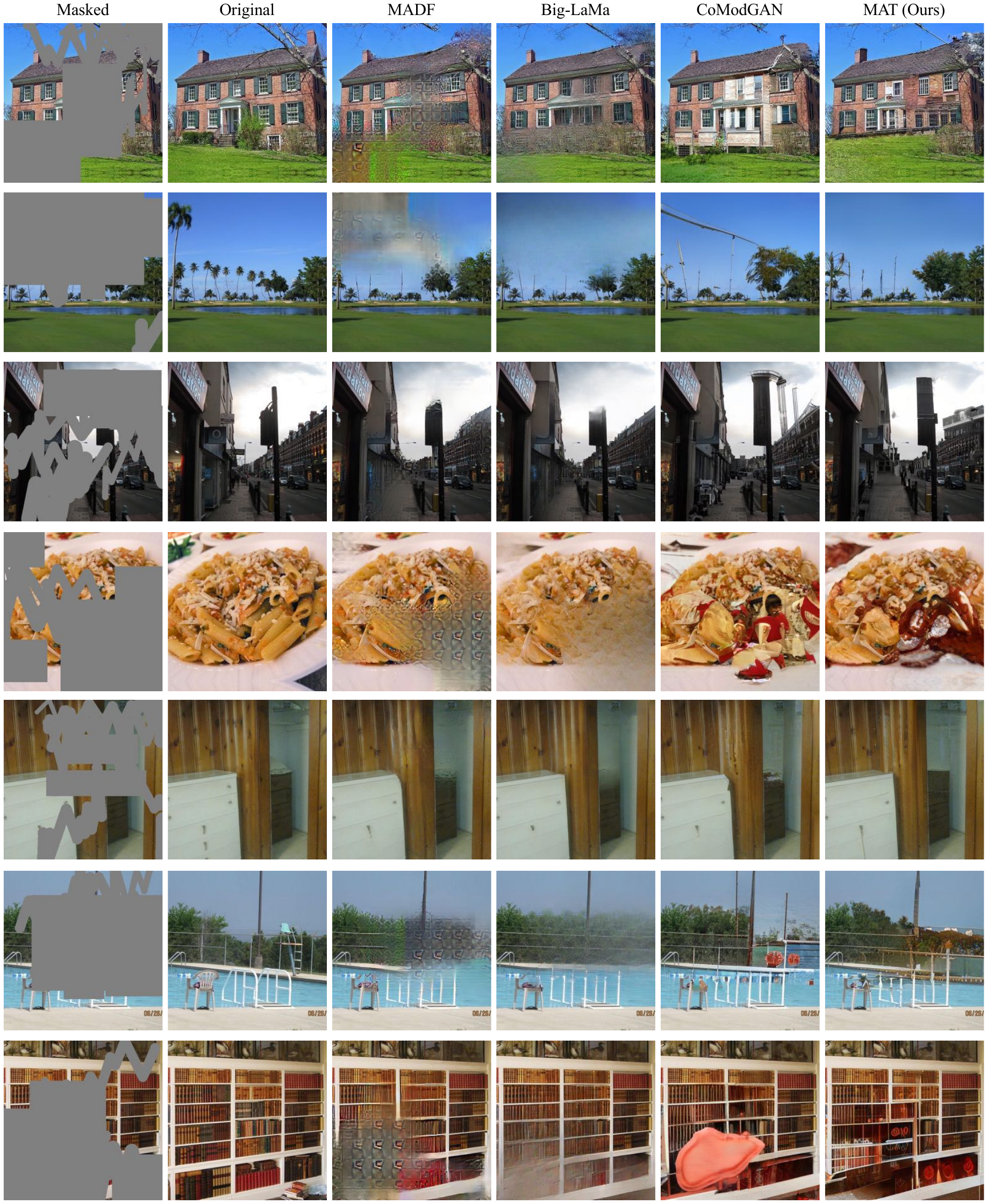}
		\end{center}
		\caption{Qualitative comparison ($512 \times 512$) with state-of-the-art methods on the Places dataset. Zoom in for a better view.}
		\label{fig:sota2}
	\end{figure*}

	We present more visual comparisons on the Places~\cite{zhou2017places} dataset between our MAT and other state-of-the-art methods. As shown in Fig~\ref{fig:sota1} and Fig~\ref{fig:sota2}, our method generates more photo-realistic results with few artifacts, manifesting the effectiveness of MAT. Due to potential copyright issues with CelebA-HQ~\cite{karras2018progressive}, we do not provide visual comparisons on this dataset. If necessary, you can process CelebA-HQ images with the provided code and model, or contact the authors. 
	
	\section{Licenses of Face Images}
	\label{sec:license}
	All face images used in the paper and supplementary material are from the FFHQ~\cite{karras2019style} dataset. Here we provide the detailed information on source and license.
	\begin{itemize}
		\item Face image in Fig.1 of main paper, source: \url{https://www.flickr.com/photos/v63/5876049365/}, license: CC BY-NC 2.0 (\url{https://creativecommons.org/licenses/by-nc/2.0/}).
		
		\item Face image in Fig.2 of main paper, source: \url{https://www.flickr.com/photos/tbisaacs/4089001580/}, license: CC BY 2.0 (\url{https://creativecommons.org/licenses/by/2.0/}).
		
		\item The first face image in Fig.6 of main paper, source: \url{https://www.flickr.com/photos/southlanarkshirecouncil/8341157963/},  license: CC BY-NC 2.0 (\url{https://creativecommons.org/licenses/by-nc/2.0/}).
		
		\item The second face image in Fig.6 of main paper, source: \url{https://www.flickr.com/photos/afge/34804627253/}, license: CC BY 2.0 (\url{https://creativecommons.org/licenses/by/2.0/}).
	\end{itemize}

\end{document}